%% file: colm2025_conference.tex
\documentclass{article} 
\PassOptionsToPackage{table}{xcolor}
\usepackage{colm2025_conference}

\usepackage{microtype}
\usepackage{hyperref}
\usepackage{url}
\usepackage{booktabs}
\usepackage{multirow}
\usepackage{tabularx}
\usepackage{enumitem} 

\usepackage{amssymb} %
\usepackage{amsmath} %
\usepackage{amsfonts} %
\usepackage{amsthm} %

\input{math_commands}

\usepackage{lineno}
\usepackage{titlesec}
\usepackage[most]{tcolorbox}
\usepackage{caption}

\definecolor{darkblue}{rgb}{0, 0, 0.5}
\definecolor{cleanGreen}{HTML}{15803D}    
\hypersetup{colorlinks=true, citecolor=darkblue, linkcolor=darkblue, urlcolor=darkblue}
\usepackage{graphicx}
\usepackage{subcaption}
\usepackage{colortbl}
\usepackage{amsmath}
\usepackage{siunitx}
\usepackage{pifont} 
\usepackage{wrapfig}
\usepackage{fontawesome5}   
\usepackage{hyperref}

\definecolor{richpurple}{RGB}{75,46,131}
\definecolor{beige}{RGB}{245,245,220}

\newcommand{\purplecomic}[1]{%
  {\color{richpurple}\selectfont #1}%
}
\newcommand\eat[1]{}

\newcommand{\up}[1]{\textcolor{green!55!black}{#1}}   
\newcommand{\noup}[1]{\textcolor{gray}{#1}}           

\newcommand{\blackcomic}[1]{%
  {\color{black}\selectfont #1}%
}
\renewcommand{\thefootnote}{\fnsymbol{footnote}}

\tcbuselibrary{theorems}
\newcounter{theoremcounter}
\newcounter{lemmacounter}
\newcounter{remarkcounter}
\newcounter{definitioncounter}

\newtcbtheorem[use counter=theoremcounter]{theorem}{Theorem}%
{colback=purple!5!white,colframe=richpurple!80!black,fonttitle=\bfseries}{thm}
\newtcbtheorem[use counter=lemmacounter]{lemma}{Lemma}%
{colback=purple!5!white,colframe=richpurple!80!black,fonttitle=\bfseries}{lem}
\newtcbtheorem[use counter=remarkcounter]{remark}{Remark}%
{colback=beige,colframe=brown!50!black,fonttitle=\bfseries}{rem}
\newtcbtheorem[use counter=definitioncounter]{definition}{Definition}%
{colback=purple!5!white,colframe=richpurple!80!black,fonttitle=\bfseries}{def}

\definecolor{naivered}{HTML}{B91C1C}

\title{
  \purplecomic{\textbf{RLCSD}}: 
  \blackcomic{Reinforcement Learning with Contrastive On-Policy Self-Distillation}
}

\author{\textbf{Leyi Pan}$^{1,2}$\footnotemark[1], \;\textbf{Shuchang Tao}$^{2}$, \textbf{Yunpeng Zhai}$^{2}$, \textbf{Lingzhe Zhang}$^{2,3}$, \textbf{Zhaoyang Liu}$^{2}$, \\\textbf{Bolin Ding}$^{2}$, \textbf{Aiwei Liu}$^{1\dagger}$, \textbf{Lijie Wen}$^{1\dagger}$\\
$^1$Tsinghua University, $^2$Tongyi Lab\includegraphics[height=12pt]{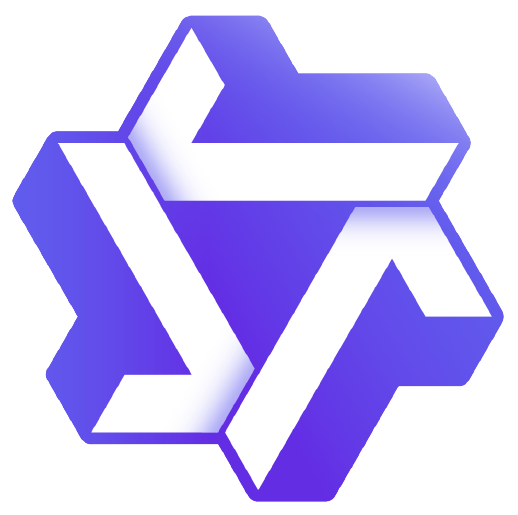}, Alibaba Group, $^3$Peking University \\\vspace{2pt}
\texttt{\small panly24@mails.tsinghua.edu.cn, liuaiwei20@gmail.com, wenlj@tsinghua.edu.cn}
}

\begin{document}


\maketitle

\begingroup
  \renewcommand\thefootnote{}%
  \footnotetext{%
    $^{*}$ This work is done during Leyi Pan's internship at Tongyi Lab, Alibaba Group.\quad
    $^{\dagger}$ Corresponding authors.%
  }%
\endgroup
\phantomsection
\vspace{-2em} 

\begin{center}
\vspace{-1.2em}
\href{https://github.com/THU-BPM/RLCSD}{\faGithub~\texttt{https://github.com/THU-BPM/RLCSD}}
\end{center}

\begin{tcolorbox}[
  colback=blue!5!white,
  colframe=richpurple!80!black,
  boxrule=1.5pt,
  arc=3mm,
  left=4mm,
  right=4mm,
  top=3mm,
  bottom=3mm,
  fonttitle=\bfseries,
  title=\textbf{Abstract}
]
On-policy self-distillation (OPSD) provides dense, token-level supervision for reasoning models by aligning a model's own distribution with that under privileged context, typically a verified solution. However, we show that the resulting distributional gap concentrates on style tokens rather than task-bearing ones, as the hinted model tends to produce shorter, more direct outputs. We term this pathology \emph{privilege-induced style drift}, which can destabilize training and shorten responses. To address this, we propose \textbf{RLCSD} (Reinforcement Learning with Contrastive on-policy Self-Distillation), which mitigates this drift by contrasting the teacher-student gap under a correct hint against that under a wrong hint, suppressing style shifts induced by hints regardless of correctness and yielding a signal more concentrated on task-bearing tokens. Experiments on Qwen3 (1.7B/4B/8B) and Olmo-3-7B-Think across mathematical and logical reasoning show that RLCSD consistently outperforms GRPO and prior OPSD methods, with additional results on agentic tasks supporting broader applicability. We further show that the contrastive principle is general: it plugs into existing OPSD methods to improve them, and its underlying insight extends to broader cross-model on-policy distillation.
\end{tcolorbox}

\begin{figure}[h!]
\vspace{-5pt}
  \centering
  \begin{subfigure}[b]{0.37\textwidth}
    \includegraphics[width=\textwidth]{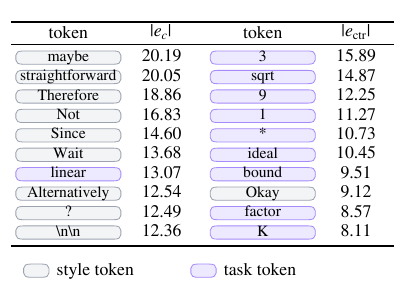}
    \caption{Top-10 tokens by $|e_c|$ and $|e_{\mathrm{ctr}}|$.}
    \label{fig:teaser-a}
  \end{subfigure}\hfill
  \begin{subfigure}[b]{0.60\textwidth}
    \includegraphics[width=\textwidth]{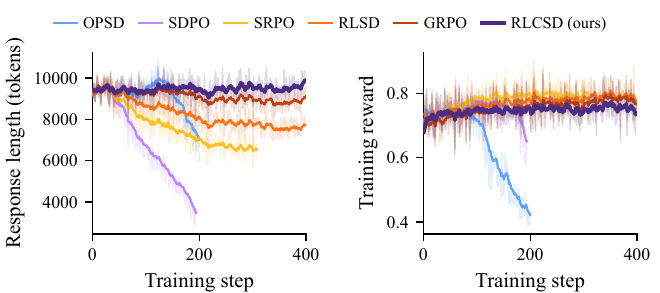}
    \caption{Response length and reward over training.}
    \label{fig:teaser-b}
  \end{subfigure}


  \begin{subfigure}[b]{\textwidth}
    \includegraphics[width=\textwidth]{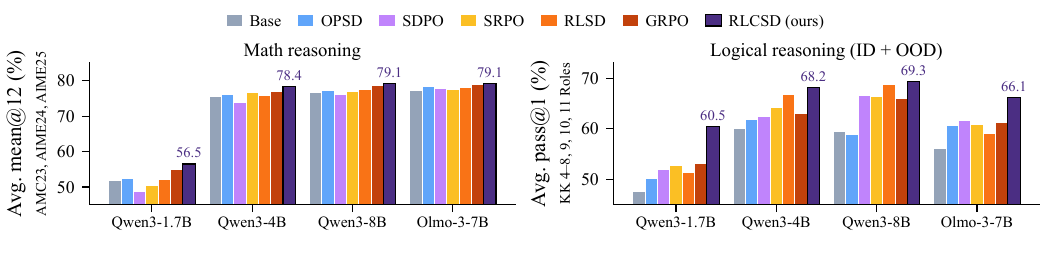}
    \caption{Performance comparison across models and benchmarks.}
    \label{fig:teaser-c}
  \end{subfigure}
  \caption{\textbf{RLCSD overview.}
\textbf{(a) Motivation.} Existing OPSD methods use $|e_c|$ as the optimization signal, which concentrates on style tokens, whereas RLCSD's contrastive signal $|e_{\mathrm{ctr}}|$ shifts the signal toward task-bearing tokens.
\textbf{(b) Response length and reward over training.} OPSD, SDPO, SRPO, and RLSD exhibit pronounced response length shrinkage during training, with OPSD and SDPO also showing training instability and sharp drops in reward. In contrast, RLCSD remains stable and maintains response length throughout training.
\textbf{(c) Performance across benchmarks.} RLCSD consistently outperforms GRPO and prior OPSD methods on mathematical and logical reasoning. }
  \label{fig:teaser}
  \vspace{-5pt}
\end{figure}

\input{sections/intro}

\input{sections/related}

\input{sections/method}

\input{sections/experiment}

\input{sections/conclusion}


\bibliography{colm2025_conference}
\bibliographystyle{colm2025_conference}

\appendix
\input{sections/appendix}

\end{document}

%% file: math_commands.tex
\usepackage{amsmath,amsfonts,bm}

\def\eqref#1{equation~\ref{#1}}

\def\1{\bm{1}}

\DeclareMathAlphabet{\mathsfit}{\encodingdefault}{\sfdefault}{m}{sl}
\SetMathAlphabet{\mathsfit}{bold}{\encodingdefault}{\sfdefault}{bx}{n}



%% file: sections/intro.tex
\section{Introduction}
\label{sec:intro}
Reinforcement learning with verifiable rewards (RLVR), exemplified by GRPO~\citep{shao2024deepseekmath}, has emerged as a dominant paradigm for training large reasoning models~\citep{guo2025deepseek,yang2025qwen3,zeng2026glm,hong2025glm,chen2026minimax,team2026kimi}. However, GRPO derives its entire learning signal from a single outcome-level reward at the end of each rollout, which becomes prohibitively sparse in long-horizon reasoning tasks where credit must be assigned across thousands of intermediate tokens. On-policy distillation (OPD)~\citep{agarwal2024policy,song2026survey,gu2024minillm} addresses this limitation by drawing dense, token-level supervision from the logits of a stronger teacher along trajectories sampled by the student itself. The on-policy formulation mitigates the exposure bias inherent in standard supervised fine-tuning, while the refinement of supervision from the trajectory level to the token level substantially accelerates convergence. Recent studies indicate that OPD attains performance comparable to or exceeding that of RLVR, and the technique has been incorporated into the post-training pipelines of several flagship large language models, including Qwen3~\citep{yang2025qwen3}, GLM-5~\citep{zeng2026glm}, the MiMo series~\citep{xiao2026mimo}, and DeepSeek-V4~\citep{deepseekai2026deepseekv4}.

The practical applicability of OPD, however, is constrained by several requirements. The method requires white-box access to the teacher's token-level logits, precluding strong closed-source teachers; the teacher and student must share an identical vocabulary, a condition rarely satisfied across model families; and serving a separate, typically larger teacher alongside the student throughout RL training imposes substantial memory and latency overhead. On-policy self-distillation (OPSD)~\citep{zhao2026self,hubotter2026reinforcement,yang2026self,he2026self,li2026unifying,jin2026unisd,kim2026rebellious,shen2026anti,shen2026generic,ke2026respecting,wang2026trace} has recently emerged as a means of circumventing these obstacles. Under this paradigm, a single model assumes both roles, with the teacher granted access to privileged context (such as a verified reference solution) unavailable to the student. 

Existing OPSD methods~\citep{zhao2026self,hubotter2026reinforcement,yang2026self} derive token-level supervision from $e_{c,t} = \log \pi_T(y_t\mid x, y_c^*, y_{<t}) - \log \pi_S(y_t\mid x, y_{<t})$, where $x$ is the query, $y$ is a student-sampled rollout, $T$ and $S$ refer to the same model under different conditioning, and $y^*_c$ is the privileged context, typically a ground truth referenced solution provided in the dataset or a correct same-group rollout. Conditioning on a verified solution does sharpen teacher confidence on task-bearing tokens, but it also induces systematic, correctness-orthogonal shifts in generative behavior: the privileged teacher favors shorter, more assertive phrasings, suppresses exploratory hedges, and reallocates probability mass toward formatting and discourse tokens. We call this parasitic component \textbf{\emph{privilege-induced style drift}}. As shown in Figure~\ref{fig:teaser-a}, stylistic tokens dominate the learning signal, diluting the signal on task-bearing tokens and leading to the pathologies illustrated in Figure~\ref{fig:teaser-b}: \textbf{premature response-length shrinkage that limits later-stage reasoning performance, sometimes accompanied by training instability}.

To address these issues, we propose \textsc{RLCSD}: \textbf{\underline{R}}einforcement \textbf{\underline{L}}earning with \textbf{\underline{C}}ontrastive on-policy \textbf{\underline{S}}elf-\textbf{\underline{D}}istillation. Rather than relying on $e_{c,t}$ alone as the learning signal, we form a contrastive estimate by conditioning the teacher on a correct and an incorrect reference solution in parallel, yielding
\begin{gather}
e_{c,t} = \log \pi_T(y_t \mid x, y_c^*, y_{<t}) - \log \pi_S(y_t \mid x, y_{<t}), \\
e_{w,t} = \log \pi_T(y_t \mid x, y_w^*, y_{<t}) - \log \pi_S(y_t \mid x, y_{<t}), \\
e_{\text{ctr},t} = e_{c,t} - e_{w,t},
\end{gather}
where $y_c^*$ and $y_w^*$ denote a correct and an incorrect reference solution, respectively. Since datasets rarely provide incorrect reference solutions, we obtain $y_w^*$ from an incorrect student rollout for the same query. The contrastive difference $e_{\text{ctr},t}$ then serves as the token-level supervision signal. When the two references are presented using the same prompt template, this subtraction can suppress stylistic components shared across the two conditioning contexts, making the remaining signal more reflective of task correctness. As illustrated in Figure~\ref{fig:teaser-a}, tokens ranked by $|e_c|$ on math problems are dominated by stylistic markers such as ``Wait'' and ``Therefore'', whereas those ranked by $|e_{\text{ctr}}|$ more often carry task-relevant information, such as mathematical content. We then use $e_{\text{ctr},t}$ to modulate the query-level advantage $A_\mathrm{ORM}$ from GRPO at each token, rather than replace it. Additional design choices, detailed in Section~\ref{sec:method}, ensure that this modulation neither reverses the direction of $A_\mathrm{ORM}$ nor allows the token-level signal to be diluted in the aggregated loss.

We evaluate RLCSD on Qwen3~\citep{yang2025qwen3} at three scales (1.7B, 4B, 8B) and Olmo-3-7B-Think~\citep{olmo2025olmo3}, with our main evaluation covering two task families: mathematical reasoning (AMC23~\citep{amc23}, AIME24~\citep{aime24}, AIME25~\citep{aime25}) and logical reasoning (Knights \& Knaves~\citep{xie2025logic}, at both in-distribution and out-of-distribution difficulty). As shown in Figure~\ref{fig:teaser-c}, RLCSD consistently outperforms GRPO and the on-policy self-distillation baselines (OPSD~\citep{zhao2026self}, SDPO~\citep{hubotter2026reinforcement}, SRPO~\citep{li2026unifying}, RLSD~\citep{yang2026self}) across the main benchmarks. Figure~\ref{fig:teaser-b} further reveals that RLCSD maintains stable training and preserves response length throughout optimization, whereas competing OPSD baselines either destabilize or suffer length shrinkage. We additionally provide experiments on agentic tasks in Appendix~\ref{app:agentic}, further demonstrating the applicability of RLCSD beyond reasoning benchmarks.

Our contributions are summarized as follows:

\begin{itemize}
\item We characterize \emph{privilege-induced style drift}, the pathology in which conditioning the teacher on a privileged solution shifts its per-token signal toward stylistic rather than correctness-bearing tokens. To address it, we propose \textbf{RLCSD}, which is designed to suppress this drift through a symmetric contrast between correct and incorrect privileged hints, and integrates the cleaned signal into RLVR as a verifier-anchored modulation of the GRPO advantage.
\item We conduct extensive experiments on Qwen3 at three scales (1.7B/4B/8B) and Olmo-3-7B-Think, showing that RLCSD \textbf{consistently outperforms} GRPO and prior on-policy self-distillation methods on mathematical and logical reasoning while maintaining stable training and response length. Additional results on agentic tasks (Appendix~\ref{app:agentic}) further demonstrate its broader applicability.
\item Extensive ablation studies show that \textbf{the contrastive principle is general}: it can be plugged into existing on-policy self-distillation methods to improve them, and our analysis extends the insight to the broader on-policy distillation setting.
\end{itemize}

%% file: sections/related.tex
\section{Related Work}
\label{sec:related}

\subsection{RLVR and On-Policy Distillation} 
Reinforcement learning with verifiable rewards (RLVR)~\citep{shao2024deepseekmath,yu2026dapo,liu2025understanding,zheng2025group,pan2025d} has emerged in recent years as a central post-training paradigm for large language models, achieving substantial gains on reasoning-intensive benchmarks. Representative algorithms such as GRPO~\citep{shao2024deepseekmath} dispense with the critic network of classical actor-critic methods and instead estimate advantages by normalizing rewards across a group of rollouts sampled from the same query. The reward itself is supplied by an external verifier and is therefore noise-free on the tasks it covers. The granularity of supervision, however, is at the level of the entire response: a single scalar advantage is broadcast uniformly to every token in the rollout, providing no fine-grained credit assignment to the intermediate reasoning steps that determine correctness. 

On-policy distillation (OPD)~\citep{gu2024minillm,agarwal2024policy} has emerged in response to this limitation. By computing a per-token distribution gap between a teacher and the student along the student's own sampled trajectories, OPD yields a dense, token-level learning signal that complements the sparse trajectory-level reward of RLVR. The dominant choice for measuring this gap is the reverse-KL divergence between teacher and student, giving rise to the canonical OPD objective:
\begin{equation}
\mathcal{L}_{\text{OPD}}(\theta) = \mathbb{E}_{y \sim \pi_S} \left[ \frac{1}{|y|} \sum_{t=1}^{|y|} \mathrm{KL}\left( \pi_S(\cdot \mid x, y_{<t}) \,\|\, \mathrm{sg}[\pi_T(\cdot \mid x, y_{<t})] \right) \right],
\end{equation}
where $\mathrm{sg}[\cdot]$ denotes the stop-gradient operator and $|y|$ is the length of the rollout in tokens. It has been shown that this objective is equivalent, in the policy-gradient sense, to REINFORCE~\citep{williams1992simple} with per-token reward $r_t=\log \pi_T(y_t\mid x,y_{<t}) - \log \pi_S(y_t\mid x,y_{<t})$ ~\citep{gu2024minillm,thinkingmachines2025opd}. 

A line of work has explored alternative divergence choices to address the trade-off between mode coverage and mode seeking inherent in reverse-KL: forward-KL offers stronger mode coverage at the cost of mass-spreading, while Jensen-Shannon Divergence~\citep{agarwal2024policy} and skew KL~\citep{ko2024distillm} formulations interpolate between the two extremes. Other extensions consider OPD under black-box access to the teacher~\citep{ye2025black}, and propose training-efficient variants that reduce the per-step distillation cost~\citep{wu2026lightning}. At industrial scale, MiMo-V2-Flash~\citep{xiao2026mimo} integrates the dense token-level advantage $\log \pi_T(y_t\mid x,y_{<t}) - \log \pi_S(y_t\mid x,y_{<t})$ from OPD with the outcome-based verifier reward $A_\mathrm{ORM}$ of RLVR in its post-training pipeline. More recent investigations~\citep{li2026rethinking,fu2026revisiting,zhu2026many} have shifted attention from objective design to the conditions under which OPD itself succeeds: the teacher and student should share compatible reasoning styles, the teacher must possess capabilities genuinely absent from the student, response length must fall within a productive regime, and approximations such as top-$k$ truncation should preserve unbiasedness of the gradient estimator.

\subsection{On-Policy Self-Distillation}
Despite the rapid progress of on-policy distillation, the paradigm faces a series of practical obstacles. First, most OPD methods require white-box access to the teacher's token-level logits, which precludes the use of strong closed-source models as teachers. Second, the teacher and student must share an identical vocabulary, a condition rarely satisfied across model families and one that restricts OPD in practice to the distillation of smaller models from larger counterparts within the same series. Third, serving a separate teacher model throughout RL training imposes substantial memory and latency overhead. To address these limitations, on-policy self-distillation (OPSD) has emerged~\citep{zhao2026self,han2026adaptive,yu2026preference,hao2026self,heakl2026cepo}. In OPSD, the teacher and the student are the same model; the teacher is simply granted access to privileged context unavailable to the student, such as a ground-truth solution or terminal feedback from the environment. The standard OPSD objective takes the following dense distillation form:
\begin{equation}
\mathcal{L}_{\text{OPSD}}(\theta) = \mathbb{E}_{y \sim \pi_S} \left[ \frac{1}{|y|} \sum_{t=1}^{|y|} \mathrm{KL}\left( \pi_S(\cdot \mid x, y_{<t}) \,\|\, \mathrm{sg}[\pi_T(\cdot \mid x, y_c^*, y_{<t})] \right) \right],
\end{equation}
where $T$ and $S$ denotes the same model, and $y_c^*$ is the privileged context. Inherited from OPD, the OPSD objective is equivalent to reinforcement learning with per-token advantage $A_{t} = \log \pi_T(y_t \mid x, y_c^*, y_{<t}) - \log \pi_S(y_t \mid x, y_{<t})$. 

Recent work has explored different instantiations of the OPSD framework, most of which adopt \emph{dense} per-token distillation. OPSD~\citep{zhao2026self} performs dense distillation using the canonical forward-KL form. While it observes the dominance of stylistic tokens in the per-token signal, it addresses this only through an ad-hoc per-token pointwise KL-clipping mechanism, which fails to resolve the issue at its root and leaves training unstable. SDPO~\citep{hubotter2026reinforcement} likewise uses dense distillation, but replaces forward KL with the Jensen--Shannon divergence to balance mode-seeking and mode-covering behavior in the per-token gap. SRPO~\citep{li2026unifying} observes that the privileged context contributes little additional signal on already-correct rollouts, and proposes a sample-level routing mechanism in which correct samples are optimized via GRPO while incorrect ones are optimized via SDPO's dense objective. In contrast to all of the above, RLSD~\citep{yang2026self} departs from dense distillation entirely: rather than matching full token-level distributions, it relies on a \emph{sampled-token} signal (which RLCSD also adopts) and argues that this OPSD-like signal should serve only as a modulation of $A_{\text{ORM}}$ rather than a replacement. Accordingly, it designs an algorithm in which the verifier-derived $A_{\text{ORM}}$ determines the direction of every update while the sampled-token signal modulates only its magnitude.

However, all of these methods rely exclusively on \emph{one-sided positive privileged contexts} (e.g., dataset ground-truth solutions or correct in-group rollouts) when constructing the teacher distribution, and therefore inherit the privilege-induced style drift identified earlier as a systematic confound in their token-level signal. \textbf{To address this limitation, we propose RLCSD, which removes the drift by construction through a symmetric contrastive cancellation between correct and incorrect privileged contexts}.

%% file: sections/method.tex
\section{Method}
\label{sec:method}

\subsection{Problem: Privilege-Induced Style Drift}
\label{sec:drift}

\begin{wrapfigure}{r}{0.42\linewidth}
  \vspace{-15pt}
  \centering
  \includegraphics[width=\linewidth]{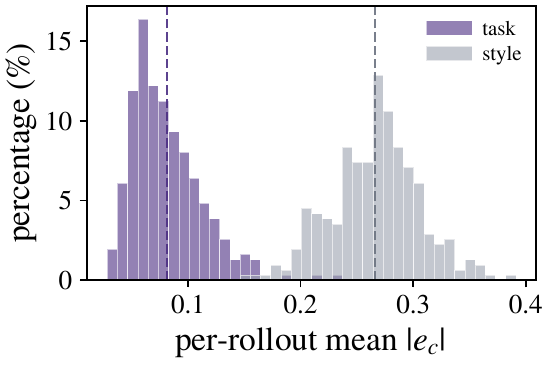}
  \caption{Per-rollout mean $|e_{c,t}|$ on style vs.\ task tokens. The
  privilege shifts most of its influence onto style tokens
  ($0.263$ vs.\ $0.083$).}
  \label{fig:ec-task-vs-style}
  \vspace{-20pt}
\end{wrapfigure}

Existing OPSD methods derive their signal from
$e_{c,t}=\log\pi_T(y_t\mid x,y^*_c,y_{<t})-\log\pi_S(y_t\mid x,y_{<t})$,
where $T$ and $S$ are the same model with and without the privileged
context $y^*_c$. Conditioning on a verified solution does sharpen the
teacher on correctness-bearing tokens, but it simultaneously induces a
systematic, correctness-orthogonal shift in generative behavior: the
privileged teacher adopts shorter, more assertive phrasings, suppresses
exploratory hedges, and reallocates probability mass toward formatting and
discourse tokens. We call this parasitic component
\emph{privilege-induced style drift}. 

\paragraph{The signal is dominated by style tokens.}
To quantify the drift, we take math reasoning as a representative case and
partition the vocabulary into a \emph{task} set of content-bearing tokens
(numerals, operators, and mathematical symbols) and a \emph{style} set of
formatting and discourse markers (e.g.\ ``Wait'', ``Therefore'',
``\textbackslash n\textbackslash n''). The full partitioning procedure is
detailed in Appendix~\ref{app:vocab-split}.
For each sampled rollout we compute the per-rollout mean of $|e_{c,t}|$
separately over its task tokens and over its style tokens, and plot the two
resulting distributions in Figure~\ref{fig:ec-task-vs-style}. The
distributions are cleanly separated: the style-token mean ($0.263$) exceeds
the task-token mean ($0.083$) by roughly $3\times$, with almost no overlap.
In other words, the privileged conditioning concentrates the overwhelming
majority of its influence on stylistic tokens while diluting the signal on
the correctness-bearing tokens that actually determine the answer. As we show
empirically in Section~\ref{sec:experiment}, this skew is not benign: in existing OPSD methods it manifests as two distinct failure modes—entropy-driven training instability, often accompanied by response-length explosion, and premature response-length shrinkage.

\subsection{Overview of RLCSD}
\label{sec:method-overview}

\begin{figure}[t]
    \centering
    \includegraphics[width=\linewidth]{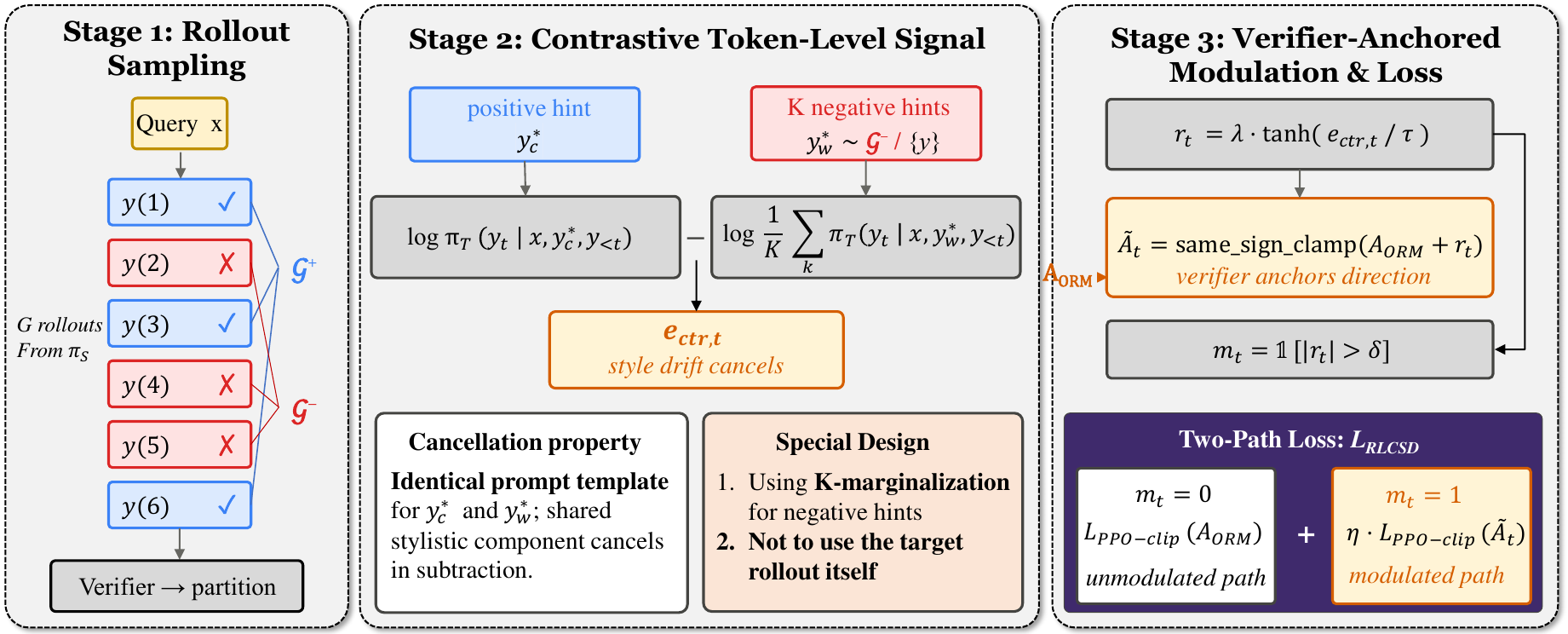}
    \caption{Overview of \textbf{RLCSD}, illustrated with the self-rollout instantiation in which the positive hint is sampled from a correct rollout.
    RLCSD more generally allows the positive privileged context to come from
    other sources, such as a dataset-provided ground-truth reasoning trace.
    \textbf{Stage 1} samples a group of rollouts and partitions them by the
    verifier into correct ($\mathcal{G}^+$) and incorrect ($\mathcal{G}^-$)
    subsets. \textbf{Stage 2} feeds a positive hint and $K$ negative hints to the teacher under an identical template; their contrast yields the token-level signal $e_{\mathrm{ctr},t}$, in which the shared privilege-induced style component is substantially suppressed. \textbf{Stage 3} converts $e_{\mathrm{ctr},t}$ into a bounded modulation $r_t$, applies it to the verifier advantage $A_{\mathrm{ORM}}$ with a sign-preserving clamp, and aggregates a two-path clipped loss.}
    \label{fig:method-pipeline}
\end{figure}

Building on the diagnosis of privilege-induced style drift in Section~\ref{sec:drift}, RLCSD constructs a clean, token-level supervision signal by contrastive cancellation across symmetrically-formatted privileged contexts and uses it to modulate the verifier-derived advantage in a GRPO objective. Figure~\ref{fig:method-pipeline} illustrates the full pipeline, which consists of three stages:
\begin{itemize}
    \item \textbf{Stage 1: Rollout Sampling and Outcome Evaluation.}
    For each query, the model samples a group of rollouts, which are evaluated
    by the verifier. These rollouts provide the outcome-level GRPO advantage and are partitioned into correct and incorrect subsets, which serve as candidate pools for rollout-derived privileged contexts.
    \item \textbf{Stage 2: Constructing the Contrastive Token-Level Signal.}
    We pair a correct privileged context with a set of incorrect privileged
    contexts under an identical prompt template. The correct context can be
    obtained either from a successful student rollout or from a
    dataset-provided ground-truth reasoning trace, while the incorrect contexts are sampled from unsuccessful student rollouts.
    The contrastive difference between the positive and the negative branches yields a per-token signal $e_{\text{ctr},t}$ on sampled tokens in which the shared stylistic component is substantially reduced, leaving a signal that is more strongly tied to task correctness.
    Moreover, we introduce two design refinements, marginalizing over $K$ negative hints and excluding the current student rollout from any rollout-derived hint pool, both detailed in \S\ref{sec:method-ectr}.
    \item \textbf{Stage 3: Modulating the Outcome Advantage and Aggregating the Loss.} We convert $e_{\text{ctr},t}$ into a bounded per-token modulation $r_t$ and apply it to $A_{\text{ORM}}$ through clamping and masking, yielding a two-path PPO-style clipped aggregated loss~\citep{schulman2017proximal}.
\end{itemize}

The remainder of this section details each stage: privileged context construction (\S\ref{sec:method-hints}), construction of the contrastive token-level signal (\S\ref{sec:method-ectr}), verifier-anchored modulation of the advantage (\S\ref{sec:method-mod}), and the final two-path loss (\S\ref{sec:method-loss}).

\subsection{Constructing Privileged Contexts}
\label{sec:method-hints}

For each query $x$, we follow the GRPO sampling protocol and draw a group of $G$ rollouts $\{y^{(g)}\}_{g=1}^G$ from the student policy $\pi_S$. A rule-based verifier assigns each rollout a binary reward: $r=1$ if the extracted answer matches the ground truth and $r=0$ otherwise\footnote{All training tasks in our experiments use binary (0/1) rewards. For math tasks the verifier extracts the $\backslash$\texttt{boxed\{\}} answer and checks equivalence; for Knights-and-Knaves it parses the assignment list.}. This partitions the group into a correct subset $\mathcal{G}^+$ and an incorrect subset $\mathcal{G}^-$. Groups for which either subset is empty are discarded; this discards no useful training signal, because a uniform-outcome group already yields zero advantage under GRPO's group-relative normalization.

\paragraph{Symmetric prompt templates for positive and negative hints.}
For each non-discarded group, we pair a \emph{positive privileged context} $y_c^*$ with one or more \emph{negative privileged contexts} drawn from $\mathcal{G}^-$. The positive context can either be sampled from a correct student rollout in $\mathcal{G}^+$ or taken from a dataset-provided ground-truth reasoning trace. Each hint is wrapped in an identical prompt template consisting of (i) the original problem statement, (ii) the reference solution's full reasoning trace, (iii) its final answer, and (iv) a continuation instruction that transitions from the reference block to the teacher's own response. The template used in our main experiments is shown below. Critically, the template makes no distinction between correct and incorrect hints: a negative hint is presented with the same ``Reference Solution'' framing and ``Correct final answer'' label as a positive one, so the teacher treats both as ground-truth references and produces distributions that differ \emph{only} in what the reference says. Because the template is byte-for-byte identical across the two branches, the privilege-induced stylistic shift is more likely to be shared between them and therefore partially cancels in the subtraction $e_{\text{ctr},t}$, leaving the signal more task-bearing in practice.

\begin{tcolorbox}[
    colback=purple!5!white,
    colframe=richpurple!80!black,
    boxrule=1pt,
    arc=2mm,
    left=3mm, right=3mm, top=2mm, bottom=2mm,
    fonttitle=\bfseries,
    title=\textbf{Teacher Privileged Prompt Template}
]
\texttt{Problem: \{problem\}}\\[4pt]
\texttt{Here is a reference solution to this problem:}\\[4pt]
\texttt{\textcolor{teal}{=== Reference Solution Begin ===}}\\[4pt]
\texttt{\{reference solution text\}}\\[2pt]
\texttt{Correct final answer: \{extracted answer\}}\\[4pt]
\texttt{\textcolor{teal}{=== Reference Solution End ===}}\\[4pt]
\texttt{After reading the reference solution above, make sure you understand the reasoning behind each step.}
\texttt{Please reason step by step, and put your final answer within} $\backslash$\texttt{boxed\{\}}\texttt{.}
\end{tcolorbox}

\subsection{Contrastive Token-Level Signal}
\label{sec:method-ectr}

\paragraph{The vanilla form of $e_{\text{ctr},t}$.}
The most vanilla instantiation of the contrastive signal pairs a correct privileged context $y_c^*$ with a single incorrect privileged context $y_w^*$, and takes the difference of the two teacher--student log-probability gaps:
\begin{equation}
    e_{\text{ctr},t}^{\textcolor{naivered}{(\text{vanilla})}}
    = e_{c,t} - e_{w,t}
    = \log \pi_T(y_t \mid x, y_c^*, y_{<t})
    - \log \pi_T(y_t \mid x, y_w^*, y_{<t}),
\end{equation}
where $y_c^*$ is either a correct student rollout sampled from $\mathcal{G}^+$ or a dataset-provided ground-truth reasoning trace, while $y_w^*$ is sampled from the incorrect rollout subset $\mathcal{G}^-$.
This vanilla form suffers from two structural problems, which motivate two corresponding refinements.

\paragraph{Refinement 1: $K$-marginalized negative hints.}
A direct use of the vanilla contrast is unstable because incorrect solutions are highly heterogeneous: for an incorrect target rollout $y$, a randomly sampled negative hint $y_w^*$ may correspond to a very different error type, in which case the teacher conditioned on $y_w^*$ does not reliably endorse the tokens of $y$. The negative branch then ceases to provide a stable counter-signal, weakening the directional consistency of $e^{(\mathrm{vanilla})}_{\mathrm{ctr},t}$ and injecting noise into training. To address this, instead of conditioning on a single incorrect rollout, we sample $K$ negative hints independently from $G^-$ and average their probabilities in the negative branch. This replaces a brittle one-to-one contrast with a more stable estimate against the error distribution represented by the group. Empirical evidence is provided in Appendix~\ref{app:negative-hint}.

\begin{wrapfigure}{r}{0.42\linewidth}
  \centering
  \includegraphics[width=\linewidth]{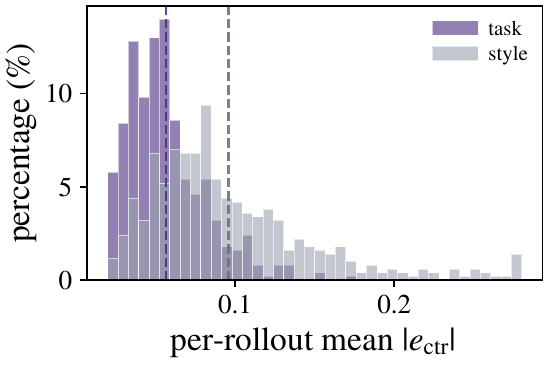}
  \caption{Task/style analysis of $|e_{\mathrm{ctr},t}|$. The reduced style--task gap indicates a cleaner signal than $|e_c|$ (Figure~\ref{fig:ec-task-vs-style}).}
  \label{fig:ctr-clean}
  \vspace{-15pt}
\end{wrapfigure}


\paragraph{Refinement 2: excluding the target rollout from the hint pool.}
When a privileged context is drawn from the rollout group, the target rollout $y$ itself may inadvertently be selected as its own hint. In this case, the teacher is given the exact response whose token probabilities it is asked to evaluate. Conditioning on $y$ induces extreme over-confidence in its tokens, making the resulting signal overly sharp and obscuring the token-level distinctions that $e_{\text{ctr},t}$ is meant to capture. We therefore \textbf{exclude $y$ from all rollout-derived hint pools}: negative hints are drawn from $\mathcal{G}^-\setminus\{y\}$, and self-generated positive hints, when used, are drawn from $\mathcal{G}^+\setminus\{y\}$. This ensures that each rollout-derived hint is a separate trajectory rather than the target rollout itself. Appendix~\ref{app:self-hint} quantifies the over-confidence effect that motivates this choice.

\paragraph{Final $e_{\mathrm{ctr},t}$ and its effect.}
Combining the two refinements, the final contrastive signal is
\begin{equation}
\small
e_{\mathrm{ctr},t}
= \log \pi_T(y_t \mid x, y_c^*, y_{<t})
- \log \frac{1}{K}\sum_{k=1}^{K}
\pi_T(y_t \mid x, y_{w,k}^*, y_{<t}),
\quad
\{y_{w,k}^*\}_{k=1}^{K}
\sim \mathcal{G}^-\!\setminus\!\{y\}.
\label{eq:final-ectr}
\end{equation}
Here, $y_c^*$ denotes the correct privileged context: it is either sampled
from $\mathcal{G}^+\setminus\{y\}$ in the self-rollout instantiation or
taken directly from a dataset-provided ground-truth reasoning trace.
Since the shared template makes the privilege-induced style component more similar across the two branches, the subtraction suppresses this shared component, leaving the signal more task-bearing. Repeating the task/style analysis of Section~\ref{sec:drift} on
$|e_{\mathrm{ctr},t}|$ (Figure~\ref{fig:ctr-clean}) confirms this: the
style--task separation seen under $|e_c|$ (Figure~\ref{fig:ec-task-vs-style})
is substantially reduced, shifting weight onto the task-bearing tokens that
determine the answer.

\subsection{Verifier-Anchored Modulation of the Advantage}
\label{sec:method-mod}

\paragraph{Outcome-level GRPO advantage.}
For each query $x$, GRPO samples a group of $G$ rollouts
$\{y^{(g)}\}_{g=1}^{G}$ from the old policy $\pi_{\theta_{\text{old}}}$.
A verifier assigns each rollout an outcome reward
$R^{(g)} \in \{0,1\}$. The outcome-level advantage is then computed
by group-relative normalization:
\begin{equation}
\label{eq:aorm}
A_{\text{ORM}}^{(g)}
=
\frac{
R^{(g)} - \frac{1}{G}\sum_{j=1}^{G} R^{(j)}
}{
\sqrt{\frac{1}{G}\sum_{j=1}^{G}
\left(R^{(j)} - \frac{1}{G}\sum_{\ell=1}^{G} R^{(\ell)}\right)^2}
+ \epsilon_{\text{adv}}
}.
\end{equation}
This scalar advantage is broadcast to all tokens in rollout $y^{(g)}$.
Thus, $A_{\text{ORM}}^{(g)}$ provides a verifier-grounded trajectory-level
update direction, but it does not distinguish which intermediate tokens
are responsible for the final outcome.

Drawing on the ideas behind Xiaomi MiMo's OPD algorithm~\citep{xiao2026mimo} and RLSD~\citep{yang2026self}, we do not treat the teacher--student log-probability gap as the sole source of learning signal. Instead, we use the verifier-derived $A_{\text{ORM}}$ as the anchor that determines the update direction, and use the contrastive token-level signal $e_{\text{ctr},t}$ only to modulate its magnitude at selected tokens. The specific procedure is as follows.

\paragraph{Soft normalization and rescaling.}
Raw $e_{\text{ctr},t}$ values can be heavy-tailed, with occasional outlier tokens producing values that would dominate the optimization. We first squash them through a hyperbolic tangent and then rescale the result by a factor $\lambda$:
\begin{equation}
\label{eq:rt}
r_t = \lambda\tanh\!\left(\frac{e_{\text{ctr},t}}{\tau}\right) \in (-\lambda, \lambda),
\end{equation}
where $\tau$ controls the soft-threshold slope and $\lambda$ sets the overall scale.

\paragraph{Token-level mask.}
Many tokens carry an $r_t$ whose absolute value is very small; these can be regarded as noise. We therefore apply a token-level mask to single out the subset of tokens carrying contrastive signal strong enough to warrant modulation, and only these tokens have their advantage modulated by $r_t$:
\begin{equation}
\label{eq:mt}
m_t = \mathbb{1}(|r_t| > \delta).
\end{equation}
Empirically, $m_t = 1$ for roughly 20\% to 30\% of tokens, consistent with the observation that a small fraction of high-influence tokens determines the outcome of long reasoning chains.

\paragraph{Sign-preserving clamp.}
For the tokens selected for modulation, we add $r_t$ to the GRPO advantage $A_{\text{ORM}}$ broadcast to token $t$ and clamp the result so that it retains the sign of $A_{\text{ORM}}$:
\begin{equation}
\label{eq:samesign}
\tilde{A}_t =
\begin{cases}
\max(0, \, A_{\text{ORM}} + r_t) & \text{if } A_{\text{ORM}} \ge 0, \\
\min(0, \, A_{\text{ORM}} + r_t) & \text{if } A_{\text{ORM}} < 0.
\end{cases}
\end{equation}
This safeguard prevents the contrastive modulation from reversing the verifier-determined direction at any individual token.

\subsection{Two-Path Loss Aggregation}
\label{sec:method-loss}

We write the RLCSD optimization target as a PPO-style clipped surrogate
objective to be maximized; in implementation, we minimize its negative
as the training loss. The objective contains two paths: an unmodulated
path for tokens with $m_t=0$, which uses the original outcome-level
advantage $A_{\text{ORM}}$, and a modulated path for tokens with $m_t=1$,
which uses the sign-preserving modulated advantage $\tilde{A}_t$.

For rollout $y^{(g)}$, let
\begin{equation}
\rho_{g,t}(\theta)
=
\frac{
\pi_\theta(y^{(g)}_t \mid x, y^{(g)}_{<t})
}{
\pi_{\theta_{\text{old}}}(y^{(g)}_t \mid x, y^{(g)}_{<t})
}
\end{equation}
denote the per-token importance ratio. For a token with advantage $A$,
we define its clipped surrogate contribution as
\begin{equation}
\label{eq:clip-surrogate}
\psi_{g,t}(A; \theta)
=
\min\!\Big(
\rho_{g,t}(\theta) A,\;
\operatorname{clip}\!\big(\rho_{g,t}(\theta), 1-\epsilon, 1+\epsilon\big) A
\Big).
\end{equation}

For each rollout $y^{(g)}$, define the unmodulated and modulated token
sets as
\begin{equation}
\mathcal{U}^{(g)} = \{t : m^{(g)}_t = 0\},
\qquad
\mathcal{M}^{(g)} = \{t : m^{(g)}_t = 1\}.
\end{equation}
The rollout-level RLCSD surrogate objective averages the two paths
independently:
\begin{equation}
\label{eq:rlcsd-rollout-obj}
J_{\text{RLCSD}}^{(g)}(\theta)
=
\frac{1}{|\mathcal{U}^{(g)}|}
\sum_{t \in \mathcal{U}^{(g)}}
\psi_{g,t}\!\left(A_{\text{ORM}}^{(g)}; \theta\right)
+
\eta \cdot
\frac{1}{|\mathcal{M}^{(g)}|}
\sum_{t \in \mathcal{M}^{(g)}}
\psi_{g,t}\!\left(\tilde{A}^{(g)}_t; \theta\right),
\end{equation}
where $\eta$ controls the relative weight of the modulated path.
In the rare case that either token set is empty, the corresponding
path is omitted from the rollout-level objective.

Finally, the full query-level objective averages over the sampled group,
and the population objective averages over training queries:
\begin{equation}
\label{eq:rlcsd-full-obj}
J_{\text{RLCSD}}(\theta)
=
\mathbb{E}_{x \sim \mathcal{D}}
\left[
\frac{1}{G}
\sum_{g=1}^{G}
J_{\text{RLCSD}}^{(g)}(\theta)
\right].
\end{equation}
The training loss minimized in implementation is the negative of this
maximization objective:
\begin{equation}
\label{eq:rlcsd-loss}
\mathcal{L}_{\text{RLCSD}}(\theta)
=
- J_{\text{RLCSD}}(\theta).
\end{equation}

The choice to normalize the two paths independently within each rollout, rather than taking a single global mean over all tokens, is essential for preserving the influence of the contrastively modulated tokens. In practice, only about 20\%--30\% of tokens satisfy $m_t=1$; under a single global average, the modulated path would therefore contribute only in proportion to this small fraction and be easily overwhelmed by the unmodulated tokens. As a result, the token-level contrastive signal would be diluted, and the objective would drift toward plain GRPO, losing the fine-grained credit assignment that motivates RLCSD. Independent normalization avoids this dilution: it decouples the strength of the modulated path from the masked-token ratio $|\mathcal{M}^{(g)}|/|y^{(g)}|$, and makes $\eta$ the sole, explicit control over how much the contrastive signal contributes to training.

%% file: sections/experiment.tex
\section{Experiments}
\label{sec:experiment}
We conduct comprehensive experiments to answer the following research questions:

\begin{itemize}[leftmargin=1.5em]
    \item \textbf{RQ1:} How does RLCSD compare with GRPO and other OPSD baselines in terms of task performance and training stability? (\S\ref{sec:exp-main})
    \item \textbf{RQ2:} Can the proposed contrastive-hint signal improve not only RLCSD itself but also other OPSD baselines as a general-purpose component? (\S\ref{sec:exp-generality})
    \item \textbf{RQ3:} How important are the individual design choices in RLCSD, including reference-hint selection, integration with $A_{\text{ORM}}$, and the two-path loss aggregation? (\S\ref{sec:exp-ablation})
    \item \textbf{RQ4:} What broader insights does RLCSD offer for on-policy distillation (OPD)? (\S\ref{sec:exp-insights})
\end{itemize}

\subsection{Experimental Setup}
\label{sec:exp-setup}

\paragraph{Models and Datasets.} We experiment with four reasoning models, \textbf{all with thinking mode enabled during both training and validation}: Qwen3-1.7B, Qwen3-4B, Qwen3-8B~\citep{yang2025qwen3}, and Olmo-3-7B~\citep{olmo2025olmo3}. We train on two categories of tasks. (1) \textbf{Math reasoning}, for which the training data is drawn from DeepMath-103K~\citep{he2025deepmath}: we use difficulty levels 5--7 for the 1.7B model, levels 6--8 for the 4B model, and levels 7--10 for the 8B model. The test sets are AMC23~\citep{amc23}, AIME24~\citep{aime24}, and AIME25~\citep{aime25}. (2) \textbf{Logical reasoning}, for which we train on Knight-and-Knaves~\citep{xie2025logic} with 4--8 roles. The test set is also Knight-and-Knaves, comprising an in-domain test (4--8 roles, 500 questions) and an out-of-domain test (9, 10, and 11 roles, 100 questions each). Additional experiments on agentic tasks (ALFWorld~\citep{shridhar2020alfworld} and Webshop~\citep{yao2022webshop}) are provided in Appendix~\ref{app:agentic}.

\paragraph{Baselines.} We compare RLCSD against (1) \textbf{GRPO}~\citep{shao2024deepseekmath}, group relative policy optimization with binary outcome rewards verified against ground-truth answers; and the following on-policy self-distillation baselines: (2) \textbf{OPSD}~\citep{zhao2026self}, dense-logit distillation using forward KL with a per-token pointwise KL-clipping mechanism; (3) \textbf{SDPO}~\citep{hubotter2026reinforcement}, dense-logit distillation using Jensen--Shannon divergence; (4) \textbf{SRPO}~\citep{li2026unifying}, which routes successful samples to GRPO and failed samples to SDPO; and (5) \textbf{RLSD}~\citep{yang2026self}, which uses the teacher--student probability gap on sampled tokens to modulate $A_\text{ORM}$, where $A_\text{ORM}$ determines the update direction and the probability gap determines its magnitude.

\paragraph{Implementation Details.} All experiments are conducted on 8 H20 GPUs using full-parameter training. We use a learning rate of $1\times 10^{-6}$ for all tasks. For RLCSD, we keep the teacher model's weights fixed throughout training. The remaining hyperparameters are set to $K=4$, $\tau=1.3$, $\lambda=0.5$, $\delta=0.02$, and $\eta=0.5$. All prior OPSD baselines and RLCSD use dataset-provided ground-truth solutions as the correct privileged contexts. For RLCSD, negative contexts are sampled from incorrect student rollouts for the same query. All contexts include both the reasoning trace and the final answer. We use identical training and evaluation settings across methods, with maximum generation lengths of $16384$ tokens during training and $38912$ tokens during validation. Further experimental details are provided in Appendix~\ref{app:implement-detail}.

\subsection{Main Results}
\label{sec:exp-main}
\input{tabs/main-tab}

\begin{figure*}[t]
    \centering
    \begin{subfigure}{0.24\textwidth}
        \includegraphics[width=\linewidth]{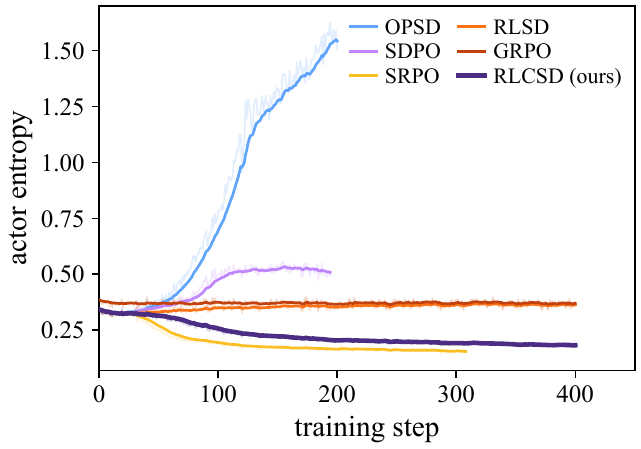}
        \caption{Math: entropy}
        \label{fig:math_entropy}
    \end{subfigure}
    \hfill
    \begin{subfigure}{0.24\textwidth}
        \includegraphics[width=\linewidth]{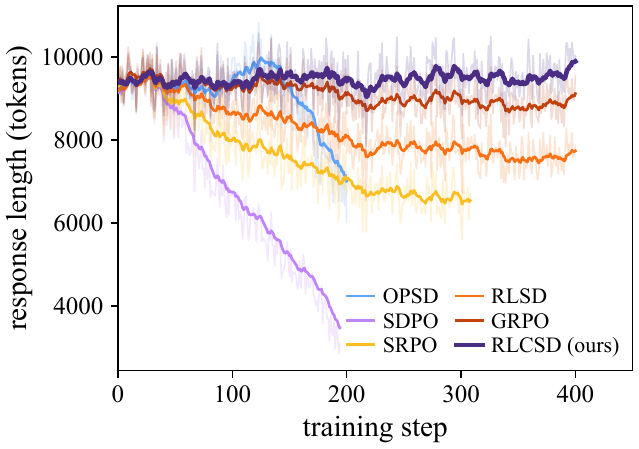}
        \caption{Math: length}
        \label{fig:math_length}
    \end{subfigure}
    \hfill
    \begin{subfigure}{0.24\textwidth}
        \includegraphics[width=\linewidth]{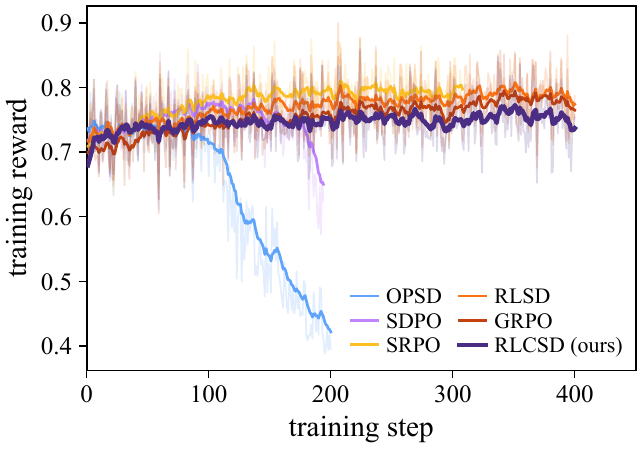}
        \caption{Math: reward}
        \label{fig:math_reward}
    \end{subfigure}
    \hfill
    \begin{subfigure}{0.24\textwidth}
        \includegraphics[width=\linewidth]{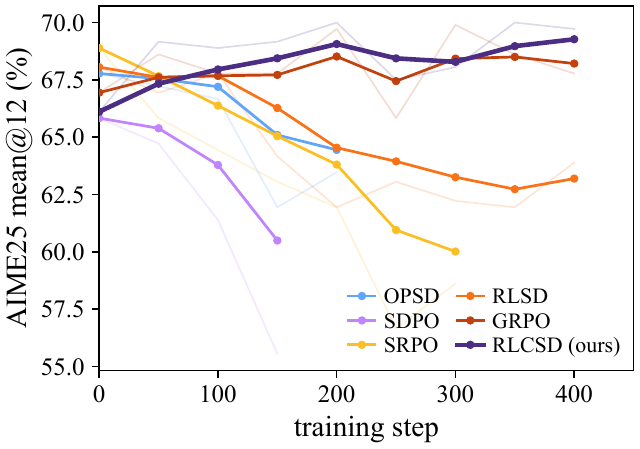}
        \caption{Math: accuracy}
        \label{fig:math_accuracy}
    \end{subfigure}

    \vskip 0.5em

    \begin{subfigure}{0.24\textwidth}
        \includegraphics[width=\linewidth]{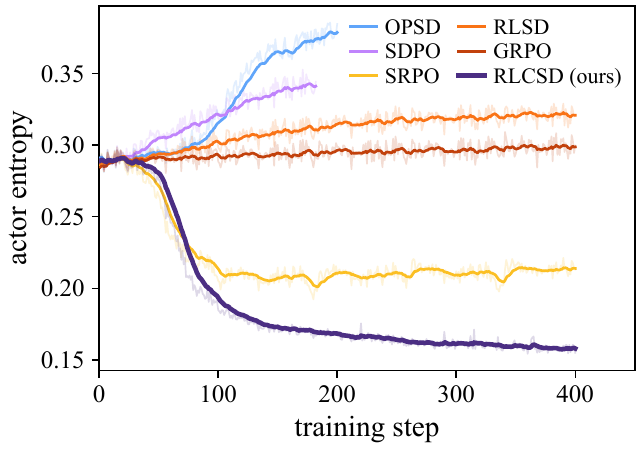}
        \caption{K\&K: entropy}
        \label{fig:KK_entropy}
    \end{subfigure}
    \hfill
    \begin{subfigure}{0.24\textwidth}
        \includegraphics[width=\linewidth]{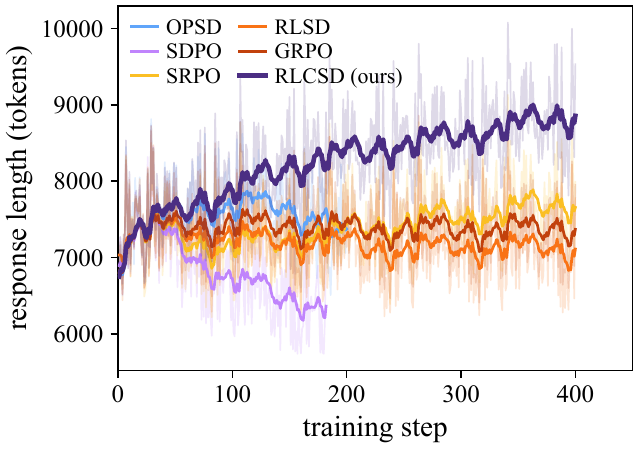}
        \caption{K\&K: length}
        \label{fig:KK_length}
    \end{subfigure}
    \hfill
    \begin{subfigure}{0.24\textwidth}
        \includegraphics[width=\linewidth]{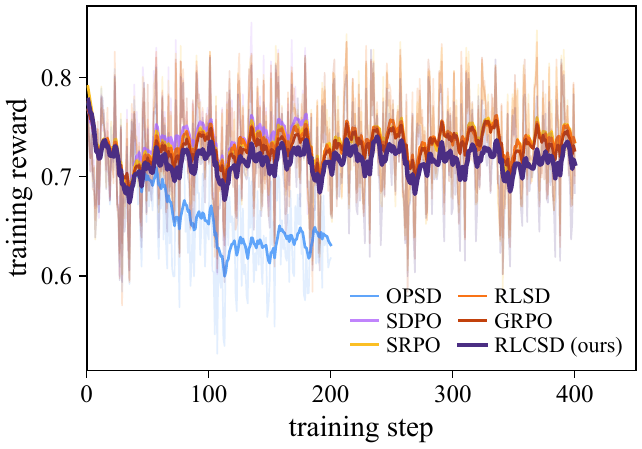}
        \caption{K\&K: reward}
        \label{fig:KK_reward}
    \end{subfigure}
    \hfill
    \begin{subfigure}{0.24\textwidth}
        \includegraphics[width=\linewidth]{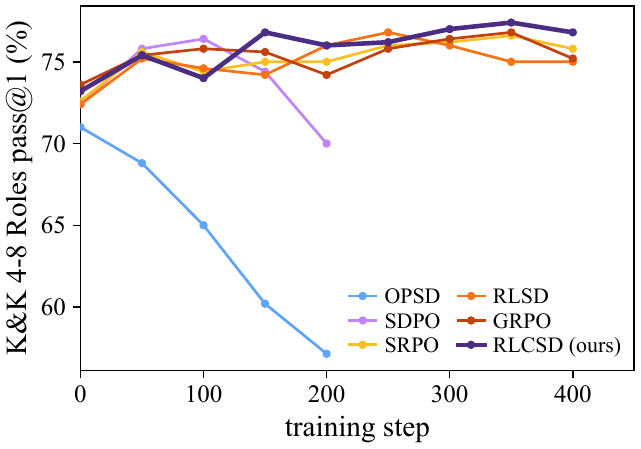}
        \caption{K\&K: accuracy}
        \label{fig:KK_accuracy}
    \end{subfigure}

\caption{Training dynamics of Qwen3-8B on math reasoning (top) and Knight-and-Knaves (bottom), comparing entropy, response length, training reward, and validation performance. Existing on-policy self-distillation methods exhibit premature response-length shrinkage, sometimes accompanied by rising entropy, training instability, and declining validation performance. In contrast, RLCSD maintains stable training dynamics, preserves response length on math, and gradually extends it on logical reasoning, while sustaining strong validation performance on both tasks.}
    \label{fig:training-curves}
\end{figure*}

\paragraph{Performance Comparison of RLCSD against Other Baselines.}
As shown in Table~\ref{tab:main}, RLCSD achieves the highest average performance on both math and logical reasoning across all four model scales, outperforming GRPO and prior OPSD methods. It achieves or ties the best result on nearly every individual benchmark. The average gains over the Base model are $+4.6/+12.9$ on math/logical reasoning for Qwen3-1.7B, $+2.9/+8.1$ for Qwen3-4B, $+2.5/+9.7$ for Qwen3-8B, and $+1.8/+9.9$ for Olmo-3-7B. Improvements are particularly pronounced on the harder OOD logical-reasoning settings, with gains ranging from $+9.0$ to $+18.0$ points across the 9--11 Roles benchmarks. These results suggest that the contrastive token-level signal supports generalization beyond the training distribution.

\paragraph{Failure Mode Analysis of Other On-policy Self-distillation Methods.}
Figure~\ref{fig:training-curves} shows that existing on-policy self-distillation methods exhibit \textbf{premature response-length shrinkage, sometimes accompanied by training instability}. 

\begin{itemize}
    \item On math reasoning, OPSD, SDPO, SRPO, and RLSD all show declining response lengths during training (Figure~\ref{fig:math_length}), alongside deteriorating or limited later-stage validation performance (Figure~\ref{fig:math_accuracy}). Meanwhile, OPSD and SDPO additionally exhibit rising entropy and unstable reward trajectories, showing a particularly sharp reward drop (Figures~\ref{fig:math_entropy} and~\ref{fig:math_reward}).
    \item On logical reasoning, response-length shrinkage is most evident for SDPO, while the other baselines generally plateau at shorter lengths than RLCSD (Figure~\ref{fig:KK_length}). OPSD and SDPO also exhibit increasing entropy and substantial declines in validation accuracy (Figures~\ref{fig:KK_entropy} and~\ref{fig:KK_accuracy}), with OPSD showing a sustained drop in training reward (Figure~\ref{fig:KK_reward}). 
\end{itemize}

\paragraph{Stable Dynamics of RLCSD.}
In contrast, RLCSD maintains stable training dynamics throughout optimization, keeping both policy entropy and response length stable (Figures~\ref{fig:math_entropy},~\ref{fig:math_length},~\ref{fig:KK_entropy}, and~\ref{fig:KK_length}), on par with GRPO. The key difference is that RLCSD augments GRPO's verifier-derived outcome signal with a dense token-level contrastive modulation, yielding finer-grained credit assignment without sacrificing optimization stability. As a result, RLCSD retains the same stable optimization behavior as GRPO while converging to consistently better validation performance, as shown in Figures~\ref{fig:math_accuracy} and~\ref{fig:KK_accuracy}.

\input{tabs/contrastive}

\begin{figure}[t]
  \centering
  \begin{minipage}[t]{0.49\textwidth}
    \centering
    \begin{subfigure}[t]{0.49\linewidth}
      \includegraphics[width=\linewidth]{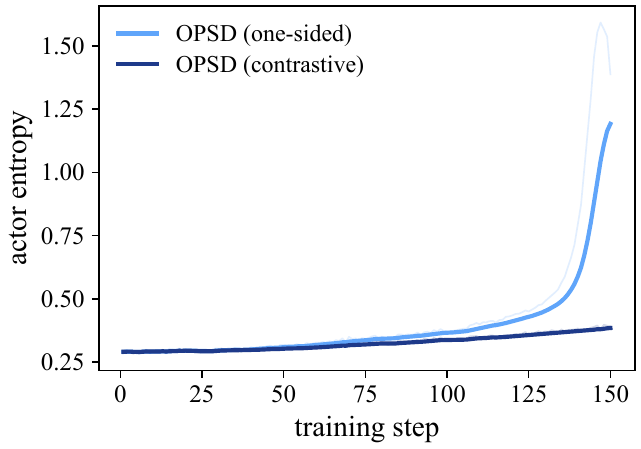}
      \caption{Entropy}
      \label{fig:opsd-entropy}
    \end{subfigure}
    \hfill
    \begin{subfigure}[t]{0.49\linewidth}
      \includegraphics[width=\linewidth]{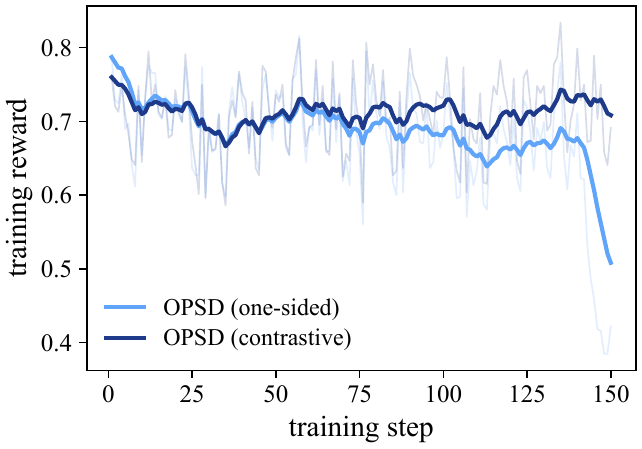}
      \caption{Training reward}
      \label{fig:opsd-reward}
    \end{subfigure}
    \addtocounter{figure}{-1}
    \captionof{figure}{Effect of the contrast mechanism on OPSD for the K\&K task (Qwen3-4B).  Without contrast, the actor entropy explodes and training
      destabilizes; adding contrast keeps the entropy bounded~(a) and yields a
      stable training reward~(b).}
    \label{fig:contrast_opsd}
  \end{minipage}
  \hfill
  \begin{minipage}[t]{0.49\textwidth}
    \centering
    \stepcounter{figure}
    \setcounter{subfigure}{0}
    \begin{subfigure}[t]{0.49\linewidth}
      \includegraphics[width=\linewidth]{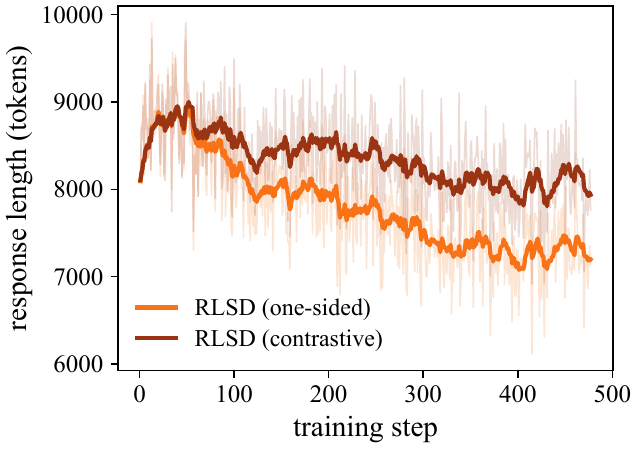}
      \caption{RLSD}
      \label{fig:rlsd-len}
    \end{subfigure}
    \hfill
    \begin{subfigure}[t]{0.49\linewidth}
      \includegraphics[width=\linewidth]{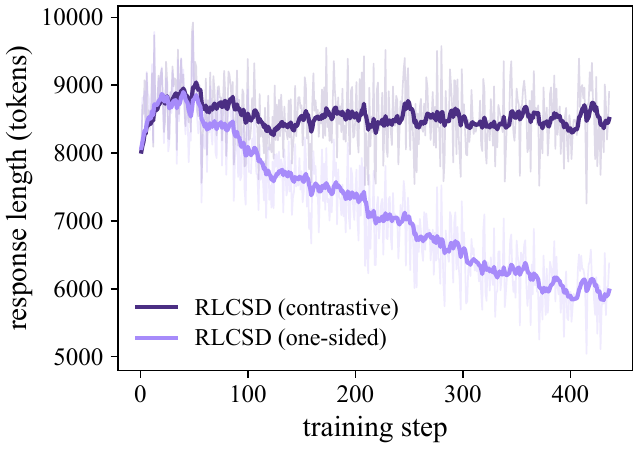}
      \caption{RLCSD}
      \label{fig:rlcsd-len}
    \end{subfigure}
    \addtocounter{figure}{-1}
    \captionof{figure}{Effect of the contrast mechanism on response length for
      math tasks (Qwen3-4B). contrast mitigates premature response-length shrinkage for both RLSD~(a) and
      one-sided RLCSD ablation~(b), preserving longer reasoning traces that benefit late-stage
      training.}
    \label{fig:contrast_length}
  \end{minipage}
\end{figure}

\subsection{Contrastive Privileged Contexts as a General-Purpose Component}
\label{sec:exp-generality}

A central claim of this work is that the contrastive construction is not specific to RLCSD, but serves as a general principle for purifying privileged token-level signals. To test this claim, we apply it to two representative on-policy self-distillation methods from different families: OPSD~\citep{zhao2026self}, a dense full-distribution distillation method, and RLSD~\citep{yang2026self}, an advantage-modulation method. For each method, we keep the underlying optimization unchanged and compare two privileged-context constructions: \textbf{(1) one-sided}, which uses the dataset-provided ground-truth reasoning trace as the positive context; and \textbf{(2) contrastive}, which contrasts the signal induced by this positive context with that induced by an incorrect hint sampled from student rollouts for the same query.

\paragraph{How the contrast enters each method.}
The two method families incorporate the contrast differently. For advantage-modulation methods (RLSD and RLCSD), the contrast acts at the scalar level: the per-token signal
\[
e_{\mathrm{ctr},t} = \log \pi_T(y_t\mid x, y_c^*, y_{<t}) - \log \frac{1}{K}\sum_k \pi_T(y_t\mid x, y_{w,k}^*, y_{<t})
\]
replaces the one-sided $e_{c,t}$ as the modulation of $A_{\mathrm{ORM}}$. OPSD, by contrast, incorporates the signal at the \emph{distribution} level. Following a classifier-free-guidance formulation~\citep{ho2022classifier}, we replace the OPSD target $\pi_T(\cdot\mid x, y_c^*, y_{<t})$ with a contrastively amplified target:
\begin{equation}
q_{\text{target}} = \mathrm{softmax}\big((1+\alpha)\,\ell_c - \alpha\,\ell_w\big),
\end{equation}
where $\ell_c$ and $\ell_w$ are the teacher logits under the correct and incorrect hints, respectively, and $\alpha\ge 0$ controls the guidance strength ($\alpha=0$ recovers vanilla OPSD). The per-token loss is further gated by a soft mask
\begin{equation}
w_t \;=\; \frac{D_t}{D_t + \tau}, \qquad D_t \;=\; \mathrm{KL}\!\left(\pi_T(\cdot\mid x, y_w^*, y_{<t})\,\big\|\,\pi_T(\cdot\mid x, y_c^*, y_{<t})\right),
\end{equation}
with $\tau\!\ge\!0$ ($\tau\!=\!0$ disables the mask), and the final objective is $\sum_t w_t\,\mathrm{KL}(q_{\text{target},t}\,\|\,\pi_{\theta}(\cdot\mid x, y_{<t}))$.

\paragraph{Results.}

\textbf{(1) The contrastive signal improves performance across different methods.} As shown in Table~\ref{tab:contrast-generality}, the contrastive construction consistently improves the math and K\&K averages across all three methods. For OPSD, it yields gains of $1.0$ and $0.9$ points on math and K\&K, respectively. For RLSD, it improves the math average by $1.4$ points, including a $2.8$-point gain on AIME24, with a marginal gain of $0.1$ points on K\&K. RLCSD shows the largest improvements, gaining $3.6$ points on math and $6.7$ points on K\&K over its one-sided counterpart. Together, these results demonstrate that the contrastive construction benefits multiple methods, supporting its applicability as a plug-in component beyond RLCSD.

\textbf{(2) The contrastive signal mitigates premature response-length shrinkage and training instability.} On K\&K, adding contrast to OPSD curbs entropy growth and stabilizes training reward (Figure~\ref{fig:contrast_opsd}). On math, it mitigates premature response-length shrinkage in RLSD (Figure~\ref{fig:rlsd-len}) and RLCSD (Figure~\ref{fig:rlcsd-len}), whose one-sided ablation exhibits a pronounced decline in response length. These observations are consistent with the failure patterns identified in Section~\ref{sec:exp-main} and suggest that contrasting correct and incorrect privileged contexts helps preserve useful token-level supervision across different optimization formulations.

\subsection{Ablation of Key Design Choices}
\label{sec:exp-ablation}

\input{tabs/ablation}

We ablate the key design choices of RLCSD introduced in Section~\ref{sec:method}, focusing on two aspects: how reference hints are selected for the privileged context, and how the contrastive token-level signal is incorporated into the final loss. All experiments in this section use Qwen3-4B, with results reported in Table~\ref{tab:ablation}.

\paragraph{Ablation on reference-hint selection.}
We first examine the two refinements introduced in Section~\ref{sec:method-ectr} for constructing the contrastive signal. The variant \textbf{$-$ $K$-marginal negative hint} replaces the marginalized negative branch with a single negative hint, making the contrast more sensitive to error-type mismatch between the target rollout and the sampled hint. The variant \textbf{$+$ target rollout in hint pool} allows the target rollout itself to serve as a hint, inducing self-referential over-confidence in the teacher and obscuring the token-level distinctions that $e_{\mathrm{ctr},t}$ is intended to capture. The mechanisms behind these effects are analyzed in Section~\ref{sec:method-ectr} and Appendix~\ref{app:hint-selection}.

Both refinements contribute to performance. Removing $K$-marginalization reduces the math and K\&K averages by 1.6 and 3.8 points, respectively, while allowing the target rollout into the hint pool reduces them by 1.6 and 4.5 points. These drops are consistent with the analysis in Section~\ref{sec:method-ectr}: the former introduces noisier negative contrasts, while the latter pushes the teacher toward trivially over-confident token distributions. Together, these results support the importance of reliable hint selection for keeping the contrastive signal aligned with task correctness.

\paragraph{Ablation on the final loss formulation.}
We next ablate the two design choices that determine how the contrastive signal enters optimization. By design, $e_{\mathrm{ctr},t}$ serves as a \emph{modulation} of $A_{\mathrm{ORM}}$ under the sign-preserving clamp in Eq.~\ref{eq:samesign}, rather than replacing it. The two signals play complementary roles: $A_{\mathrm{ORM}}$ provides a verifier-grounded update direction based on the outcome reward, while $e_{\mathrm{ctr},t}$ refines its magnitude at the token level. The variant \textbf{$-$ $A_{\mathrm{ORM}}$ anchoring} removes this anchor and uses $e_{\mathrm{ctr},t}$ directly as the token-level advantage. This causes the largest degradation among the ablations, reducing the math and K\&K averages by 2.7 and 5.8 points, respectively. Although the contrastive signal is cleaner than its one-sided counterpart, it can still assign signs that disagree with the outcome-based advantage. Anchoring on $A_{\mathrm{ORM}}$ prevents these disagreements from reversing the update direction, allowing the contrastive signal to refine credit assignment without overriding the outcome reward.

Finally, the variant \textbf{$-$ two-path loss aggregation} replaces the two-path objective in Eq.~\ref{eq:rlcsd-rollout-obj} with a single global average over all tokens. Under this formulation, the modulated path contributes only in proportion to the masked-token fraction $|\mathcal{M}|/N$. Since only about 20\%--30\% of tokens satisfy the mask in practice, this dilutes the contrastive contribution and shifts the objective toward plain GRPO. Removing two-path aggregation reduces the math and K\&K averages by 2.0 and 4.8 points, respectively, supporting the importance of independent normalization for preserving the influence of the modulated tokens.

\subsection{Broader Insights for On-Policy Distillation}
\label{sec:exp-insights}

\paragraph{The same decomposition governs cross-model OPD.}
The privilege-induced style drift we identify in \emph{self}-distillation may not be specific to using the student as its own teacher. A similar decomposition into task-bearing and surface-style components may also help explain \emph{cross-model} on-policy distillation (OPD). Standard OPD uses the per-token reverse KL between teacher $T$ and student $S$ on student rollouts as the learning signal~\citep{thinkingmachines2025opd,agarwal2024policy}. Part of this $T$--$S$ gap reflects genuine differences in reasoning ability, while another part reflects stylistic differences in discourse, formatting, and thinking pattern. When the latter dominates, the gradient transfers register rather than capability, and OPD degrades into stylistic mimicry. Consistent with this view, \citet{li2026rethinking} show that a stronger teacher can fail where a weaker but stylistically aligned one succeeds. To make this intuition concrete, we analyze which tokens dominate the learning signal across teacher--student pairs.

\begin{wraptable}{r}{0.46\textwidth}
\centering
\vspace{-12pt}
\small
\setlength{\tabcolsep}{5pt}
\caption{Top-15 response tokens by per-token $\mathrm{KL}_t$ for each teacher, on Qwen3-1.7B-Base rollouts over DeepMath.}
\vspace{-8pt}
\label{tab:opd-top15}
\begin{tabular}{c l r l r}
\toprule
& \multicolumn{2}{c}{Qwen3-4B-Instruct} & \multicolumn{2}{c}{Qwen3-4B-Base-GRPO} \\
\cmidrule(lr){2-3}\cmidrule(lr){4-5}
\# & token & KL & token & KL \\
\midrule
1  & \texttt{<|endoftext|>}        & 29.2 & \texttt{3}        & 11.6 \\
2  & \texttt{ The}                 & 25.3 & \texttt{ subsets} & 10.4 \\
3  & \texttt{ \$\textbackslash}    & 23.5 & \texttt{ \textbackslash\textbackslash} & 9.6 \\
4  & \texttt{\textbackslash n\textbackslash n} & 23.0 & \texttt{,} & 9.4 \\
5  & \texttt{ five}                & 21.9 & \texttt{\}\}}     & 8.7 \\
6  & \texttt{ If}                  & 21.2 & \texttt{!)}       & 8.7 \\
7  & \texttt{ This}                & 21.0 & \texttt{M}        & 8.0 \\
8  & \texttt{Done}                 & 18.6 & \texttt{ex}       & 8.0 \\
9  & \texttt{ Thus}                & 17.4 & \texttt{(\textbackslash\textbackslash)} & 7.5 \\
10 & \texttt{qx}                   & 15.6 & \texttt{ x}       & 7.4 \\
11 & \texttt{ \char127}            & 15.6 & \texttt{=(}       & 7.4 \\
12 & \texttt{ \textbackslash(}     & 15.1 & \texttt{1}        & 7.3 \\
13 & \texttt{Name}                 & 14.7 & \texttt{\}}       & 7.3 \\
14 & \texttt{ Therefore}           & 14.2 & \texttt{c}        & 7.0 \\
15 & \texttt{Reach}                & 14.1 & \texttt{\{}       & 6.9 \\
\bottomrule
\end{tabular}
\vspace{-10pt}
\end{wraptable}

\paragraph{Setup.}
We adopt the same teacher--student configuration as \citet{li2026rethinking}. The student is Qwen3-1.7B-Base. We compare two teachers with similar benchmark accuracy but very different stylistic distance from the student: Qwen3-4B-Instruct, a heavily post-trained instruct model in non-thinking mode, and Qwen3-4B-Base-GRPO\footnote{\url{https://huggingface.co/lllyx/Qwen3-4B-Base-GRPO}}, a lightly post-trained math model obtained by zero-RL GRPO on Qwen3-4B-Base. The former is stylistically far from a base student, while the latter remains much closer in style. We sample 200 prompts from DeepMath~\citep{he2025deepmath}, draw two student rollouts for each prompt, and compute the per-token reverse KL,
\[
\mathrm{KL}_t = \mathrm{KL}\!\left( \pi_S(\cdot \mid x, y_{<t}) \,\|\, \pi_T(\cdot \mid x, y_{<t}) \right),
\]
between the student and each teacher.

\paragraph{Results.}
Table~\ref{tab:opd-top15} reveals a clear separation in which tokens dominate the learning signal. For the instruct teacher, the highest-$\mathrm{KL}_t$ positions are dominated by discourse and structural tokens, including connectives (``The'', ``If'', ``Thus''), paragraph breaks, and LaTeX delimiters. Its gradient is therefore concentrated on stylistic choices---roughly, ``how should I phrase the next step?''---rather than on mathematical content. In contrast, the GRPO teacher's highest-$\mathrm{KL}_t$ positions are dominated by digits, operators, and task-bearing content words such as ``subsets'', ``ex'', and ``=''. This is the kind of signal that directly teaches mathematics rather than register. Training results further support this interpretation (Table~\ref{tab:opd-train}): although Qwen3-4B-Instruct has the stronger standalone math performance, distilling it into Qwen3-1.7B-Base yields a worse student than distilling the stylistically closer GRPO teacher.

\paragraph{Takeaways.}
This analysis yields two implications. First, the style drift characterized for OPSD may not be unique to self-distillation: a similar task-bearing versus surface-style decomposition appears relevant to cross-model OPD, where a teacher whose KL mass falls on discourse tokens may transfer register rather than capability. Second, because OPD conditions on no privileged hint, RLCSD's contrastive cancellation does not directly apply. Instead, the per-token KL ranking offers a cheap, training-free diagnostic for teacher selection.

\begin{table}[t]
\centering
\caption{Cross-model OPD with teachers of different stylistic distance from the student. The student is Qwen3-1.7B-Base, and both runs use the same OPD training setup (DeepMath~\citep{he2025deepmath} as training set, 300 steps). Although Qwen3-4B-Instruct has stronger standalone math performance, distilling from the stylistically closer Qwen3-4B-Base-GRPO yields a better student on all three math benchmarks.}
\label{tab:opd-train}
\small
\resizebox{\textwidth}{!}{%
\begin{tabular}{lcccccccc}
\toprule
& \multicolumn{4}{c}{Teacher standalone (mean@12)} & \multicolumn{4}{c}{Distilled student (mean@12)} \\
\cmidrule(lr){2-5}\cmidrule(lr){6-9}
Teacher & AMC23 & AIME24 & AIME25 & Avg. & AMC23 & AIME24 & AIME25 & Avg. \\
\midrule
Qwen3-4B-Instruct   & \textbf{60.8} & \textbf{24.6} & \textbf{19.2} & \textbf{34.9} & 32.9 & 9.0 & 5.2 & 15.7 \\
Qwen3-4B-Base-GRPO &  56.1 & 22.3 & 19.0 & 32.5 & \textbf{39.7} & \textbf{13.3} & \textbf{9.2} & \textbf{20.7} \\
\midrule
Base student (Qwen3-1.7B-Base) & \multicolumn{4}{c}{--} & 12.0 & 1.5 & 1.7 & 5.1 \\
\bottomrule
\end{tabular}
}
\end{table}


%% file: tabs/main-tab.tex
\begin{table}[t]
\centering
\caption{Main results across model scales. Math reasoning (AMC23, AIME24, AIME25) is reported as mean@12; logical reasoning (Knight-and-Knaves) is reported as pass@1, with 4--8 Roles as the in-domain (ID) setting and 9 Roles, 10 Roles, and 11 Roles as out-of-domain (OOD) settings. \textbf{Bold} marks the best result in each block, and \underline{underline} marks the second-highest result; \textcolor{green!50!black}{green subscripts} denote the gain over the Base model.}
\label{tab:main}
\setlength{\tabcolsep}{4pt}
\resizebox{\textwidth}{!}{%
\begin{tabular}{llcccc ccccc}
\toprule
& & \multicolumn{4}{c}{\textbf{Math Reasoning} (mean@12)} & \multicolumn{5}{c}{\textbf{Logical Reasoning: Knight-and-Knaves} (pass@1)} \\
\cmidrule(lr){3-6}\cmidrule(lr){7-11}
\textbf{Model} & \textbf{Method} & AMC23 & AIME24 & AIME25 & Avg. & 4--8 Roles & 9 Roles & 10 Roles & 11 Roles & Avg. \\
\midrule

\multirow{7}{*}{Qwen3-1.7B}
 & Base  & 74.1 & 48.3 & 33.3 & 51.9 & 63.2 & 53.0 & 43.0 & 31.0 & 47.6 \\
 & GRPO  & \underline{76.6} & \underline{51.6} & \underline{37.2} & \underline{55.1} & 67.4 & \underline{59.0} & 52.0 & 34.0 & \underline{53.1} \\
 & OPSD  & 74.6 & 47.2 & 36.1 & 52.6 & 62.0 & \underline{59.0} & 51.0 & 29.0 & 50.3 \\
 & SDPO  & 73.0 & 42.2 & 31.4 & 48.9 & 63.8 & 56.0 & 52.0 & \underline{36.0} & 52.0 \\
 & SRPO  & 72.6 & 45.6 & 33.3 & 50.5 & 63.0 & \underline{59.0} & \underline{54.0} & 35.0 & 52.8 \\
 & RLSD  & 73.9 & 46.1 & 36.9 & 52.3 & \underline{68.0} & 57.0 & 49.0 & 32.0 & 51.5 \\
 \rowcolor{gray!30}
 & \textbf{RLCSD}
   & \textbf{78.2}$_{\textcolor{green!50!black}{+4.1}}$
   & \textbf{53.1}$_{\textcolor{green!50!black}{+4.8}}$
   & \textbf{38.3}$_{\textcolor{green!50!black}{+5.0}}$
   & \textbf{56.5}$_{\textcolor{green!50!black}{+4.6}}$
   & \textbf{71.8}$_{\textcolor{green!50!black}{+8.6}}$
   & \textbf{66.0}$_{\textcolor{green!50!black}{+13.0}}$
   & \textbf{61.0}$_{\textcolor{green!50!black}{+18.0}}$
   & \textbf{43.0}$_{\textcolor{green!50!black}{+12.0}}$
   & \textbf{60.5}$_{\textcolor{green!50!black}{+12.9}}$ \\

\midrule

\multirow{7}{*}{Qwen3-4B}
 & Base  & 88.6 & 72.5 & 65.3 & 75.5 & 73.2 & 67.0 & 58.0 & 42.0 & 60.1 \\
 & GRPO  & 89.1 & \underline{75.8} & 66.1 & \underline{77.0} & 75.4 & 71.0 & 61.0 & 45.0 & 63.1 \\
 & OPSD  & \underline{89.4} & 73.3 & 65.8 & 76.2 & 73.4 & 71.0 & 61.0 & 42.0 & 61.9 \\
 & SDPO  & 88.2 & 70.3 & 63.6 & 74.0 & 75.8 & 71.0 & 60.0 & 43.0 & 62.5 \\
 & SRPO  & 89.2 & 74.2 & \underline{66.7} & 76.7 & 74.6 & \textbf{76.0} & 59.0 & 47.0 & 64.2 \\
 & RLSD  & 88.9 & 72.5 & 66.1 & 75.8 & \underline{76.0} & \underline{75.0} & \underline{67.0} & \underline{49.0} & \underline{66.8} \\
 \rowcolor{gray!30}
 & \textbf{RLCSD}
   & \textbf{90.8}$_{\textcolor{green!50!black}{+2.2}}$
   & \textbf{76.4}$_{\textcolor{green!50!black}{+3.9}}$
   & \textbf{68.1}$_{\textcolor{green!50!black}{+2.8}}$
   & \textbf{78.4}$_{\textcolor{green!50!black}{+2.9}}$
   & \textbf{77.6}$_{\textcolor{green!50!black}{+4.4}}$
   & \textbf{76.0}$_{\textcolor{green!50!black}{+9.0}}$
   & \textbf{68.0}$_{\textcolor{green!50!black}{+10.0}}$
   & \textbf{51.0}$_{\textcolor{green!50!black}{+9.0}}$
   & \textbf{68.2}$_{\textcolor{green!50!black}{+8.1}}$ \\

\midrule

\multirow{7}{*}{Qwen3-8B}
 & Base  & 88.8 & 74.2 & 66.9 & 76.6 & 72.4 & 67.0 & 55.0 & 44.0 & 59.6 \\
 & GRPO  & \underline{90.1} & \underline{76.1} & \underline{69.7} & \underline{78.6} & \underline{76.8} & 75.0 & 63.0 & 49.0 & 66.0 \\
 & OPSD  & 90.0 & 74.7 & 67.2 & 77.3 & 68.8 & 65.0 & 56.0 & 46.0 & 59.0 \\
 & SDPO  & \underline{90.1} & 74.2 & 64.7 & 76.3 & 76.4 & \underline{78.0} & 66.0 & 46.0 & 66.6 \\
 & SRPO  & 89.4 & 75.6 & 65.8 & 76.9 & 76.6 & 76.0 & 63.0 & 50.0 & 66.4 \\
 & RLSD  & 89.6 & 75.6 & 67.8 & 77.7 & \underline{76.8} & \textbf{79.0} & \textbf{67.0} & \underline{52.0} & \underline{68.7} \\
 \rowcolor{gray!30}
 & \textbf{RLCSD}
   & \textbf{91.0}$_{\textcolor{green!50!black}{+2.2}}$
   & \textbf{76.4}$_{\textcolor{green!50!black}{+2.2}}$
   & \textbf{70.0}$_{\textcolor{green!50!black}{+3.1}}$
   & \textbf{79.1}$_{\textcolor{green!50!black}{+2.5}}$
   & \textbf{78.2}$_{\textcolor{green!50!black}{+5.8}}$
   & \textbf{79.0}$_{\textcolor{green!50!black}{+12.0}}$
   & \underline{66.0}$_{\textcolor{green!50!black}{+11.0}}$
   & \textbf{54.0}$_{\textcolor{green!50!black}{+10.0}}$
   & \textbf{69.3}$_{\textcolor{green!50!black}{+9.7}}$ \\

\midrule

\multirow{7}{*}{Olmo-3-7B}
 & Base  & 91.2 & 73.9 & 66.9 & 77.3 & 70.6 & 64.0 & 55.0 & 35.0 & 56.2 \\
 & GRPO  & 92.4 & \underline{75.8} & \textbf{68.9} & \underline{79.0} & \underline{73.8} & 69.0 & \underline{63.0} & 39.0 & 61.2 \\
 & OPSD  & 92.3 & 75.4 & 67.1 & 78.3 & 72.6 & \underline{70.0} & 61.0 & 39.0 & 60.7 \\
 & SDPO  & 91.8 & 74.3 & 67.2 & 77.8 & 73.4 & 68.0 & 60.0 & \underline{45.0} & \underline{61.6} \\
 & SRPO  & 91.9 & 75.2 & 65.6 & 77.6 & 73.0 & 67.0 & 62.0 & 41.0 & 60.8 \\
 & RLSD  & \underline{92.5} & 74.9 & 66.8 & 78.1 & 73.6 & 66.0 & 60.0 & 37.0 & 59.2 \\
 \rowcolor{gray!30}
 & \textbf{RLCSD}
   & \textbf{92.7}$_{\textcolor{green!50!black}{+1.5}}$
   & \textbf{76.1}$_{\textcolor{green!50!black}{+2.2}}$
   & \underline{68.6}$_{\textcolor{green!50!black}{+1.7}}$
   & \textbf{79.1}$_{\textcolor{green!50!black}{+1.8}}$
   & \textbf{75.4}$_{\textcolor{green!50!black}{+4.8}}$
   & \textbf{76.0}$_{\textcolor{green!50!black}{+12.0}}$
   & \textbf{65.0}$_{\textcolor{green!50!black}{+10.0}}$
   & \textbf{48.0}$_{\textcolor{green!50!black}{+13.0}}$
   & \textbf{66.1}$_{\textcolor{green!50!black}{+9.9}}$ \\

\bottomrule
\end{tabular}%
}
\end{table}

%% file: tabs/contrastive.tex
\begin{table}[t]
\centering
\caption{Contrastive privileged contexts as a plug-in component. For each method, we compare a one-sided construction using a correct reference solution with a contrastive construction that additionally uses an incorrect student rollout for the same query. Math reasoning (AMC23, AIME24, AIME25) is reported as mean@12; Knight-and-Knaves is reported as pass@1, with 4--8 Roles as the in-domain (ID) setting and 9--11 Roles as out-of-domain (OOD) settings. All results use Qwen3-4B; \textbf{bold} marks the best result within each method block, including ties. $\Delta$ denotes the gain of the contrastive construction over its one-sided counterpart.}
\label{tab:contrast-generality}
\setlength{\tabcolsep}{4pt}
\resizebox{\textwidth}{!}{%
\begin{tabular}{ll cccc ccccc}
\toprule
& & \multicolumn{4}{c}{\textbf{Math Reasoning} (mean@12)} & \multicolumn{5}{c}{\textbf{Logical Reasoning: Knight-and-Knaves} (pass@1)} \\
\cmidrule(lr){3-6}\cmidrule(lr){7-11}
\textbf{Method} & \textbf{Context construction} & AMC23 & AIME24 & AIME25 & Avg. & 4--8 Roles & 9 Roles & 10 Roles & 11 Roles & Avg. \\
\midrule
\multirow{3}{*}{OPSD}
  & One-sided & 89.4 & 73.3 & 65.8 & 76.2 & 73.4 & \textbf{71.0} & \textbf{61.0} & 42.0 & 61.9 \\
  & Contrastive & \textbf{89.6} & \textbf{75.6} & \textbf{66.4} & \textbf{77.2} & \textbf{74.2} & \textbf{71.0} & \textbf{61.0} & \textbf{45.0} & \textbf{62.8} \\
\cmidrule(lr){2-11}
  & $\Delta$ & \up{+0.2} & \up{+2.3} & \up{+0.6} & \up{+1.0} & \up{+0.8} & \noup{0.0} & \noup{0.0} & \up{+3.0} & \up{+0.9} \\
\midrule
\multirow{3}{*}{RLSD}
  & One-sided & 88.9 & 72.5 & 66.1 & 75.8 & 76.0 & \textbf{75.0} & \textbf{67.0} & \textbf{49.0} & 66.8 \\
  & Contrastive & \textbf{89.8} & \textbf{75.3} & \textbf{66.4} & \textbf{77.2} & \textbf{76.4} & \textbf{75.0} & \textbf{67.0} & \textbf{49.0} & \textbf{66.9} \\
\cmidrule(lr){2-11}
  & $\Delta$ & \up{+0.9} & \up{+2.8} & \up{+0.3} & \up{+1.4} & \up{+0.4} & \noup{0.0} & \noup{0.0} & \noup{0.0} & \up{+0.1} \\
\midrule
\multirow{3}{*}{RLCSD}
  & One-sided & 88.9 & 73.1 & 62.5 & 74.8 & 73.8 & 71.0 & 58.0 & 43.0 & 61.5 \\
  & Contrastive & \textbf{90.8} & \textbf{76.4} & \textbf{68.1} & \textbf{78.4} & \textbf{77.6} & \textbf{76.0} & \textbf{68.0} & \textbf{51.0} & \textbf{68.2} \\
\cmidrule(lr){2-11}
  & $\Delta$ & \up{+1.9} & \up{+3.3} & \up{+5.6} & \up{+3.6} & \up{+3.8} & \up{+5.0} & \up{+10.0} & \up{+8.0} & \up{+6.7} \\
\bottomrule
\end{tabular}%
}
\end{table}

%% file: tabs/ablation.tex
\begin{table}[t]
\centering
\caption{Ablation of key design choices in RLCSD (Qwen3-4B). Each variant removes a single component from the full method. Math reasoning (AMC23, AIME24, AIME25) is reported as mean@12; Knight-and-Knaves (pass@1) uses roles 4--8 as in-domain (ID) and roles 9/10/11 as out-of-domain (OOD). Bold marks the best result in each column.}
\label{tab:ablation}
\small
\resizebox{\textwidth}{!}{%
\begin{tabular}{lccccccccc}
\toprule
& \multicolumn{4}{c}{Math Reasoning (mean@12)} & \multicolumn{5}{c}{Logical Reasoning: Knight-and-Knaves (pass@1)} \\
\cmidrule(lr){2-5} \cmidrule(lr){6-10}
Method & AMC23 & AIME24 & AIME25 & Avg. & 4--8 Roles & 9 Roles & 10 Roles & 11 Roles & Avg. \\
\midrule
RLCSD (full) & \textbf{90.8} & \textbf{76.4} & \textbf{68.1} & \textbf{78.4} & \textbf{77.6} & \textbf{76.0} & \textbf{68.0} & \textbf{51.0} & \textbf{68.2} \\
$-$ $K$-marginal negative hint & 89.3 & 73.2 & 67.8 & 76.8 & 76.4 & 71.0 & 63.0 & 47.0 & 64.4 \\
$+$ target rollout in hint pool & 89.1 & 73.4 & 68.0 & 76.8 & 75.8 & 70.0 & 63.0 & 46.0 & 63.7 \\
$-$ $A_{\mathrm{ORM}}$ anchoring & 88.6 & 72.1 & 66.3 & 75.7 & 74.6 & 69.0 & 61.0 & 45.0 & 62.4 \\
$-$ two-path loss aggregation & 89.0 & 73.1 & 67.2 & 76.4 & 75.6 & 70.0 & 62.0 & 46.0 & 63.4 \\
\bottomrule
\end{tabular}
}
\end{table}

%% file: sections/conclusion.tex
\section{Conclusion}
\label{sec:conclusion}

We studied on-policy self-distillation through the lens of \emph{privilege-induced style drift}, a pathology in which privileged conditioning shifts the token-level learning signal toward stylistic rather than task-bearing tokens. To address this, we proposed RLCSD, which mitigates the drift by contrasting correct and incorrect privileged hints under a shared template, and integrates the resulting signal into RLVR as a verifier-anchored token-level modulation of $A_{\mathrm{ORM}}$. Experiments across Qwen3 and Olmo models on mathematical and logical reasoning, as well as agentic tasks, show that RLCSD consistently outperforms GRPO and prior OPSD methods while maintaining stable training dynamics. Beyond RLCSD itself, our results show that the contrastive principle is general: it improves existing OPSD methods and offers a broader perspective on how token-level distillation signals should be constructed and interpreted in both self-distillation and cross-model OPD.

%% file: sections/appendix.tex
\newpage

\section{Additional Analysis of Reference-Hint Selection}
\label{app:hint-selection}

This appendix expands on the two refinements introduced in
Section~\ref{sec:method-ectr}: $K$-marginalization of negative hints
(Appendix~\ref{app:negative-hint}) and excluding the target rollout from the
hint pool (Appendix~\ref{app:self-hint}). Together they detail why the vanilla
contrastive signal is unreliable and how each refinement restores a clean,
verifier-aligned signal.

\subsection{Negative-Hint Selection: Case Analysis and \texorpdfstring{$K$}{K}-Marginalization}
\label{app:negative-hint}

\paragraph{Why a single negative hint is unreliable.}
Correct solutions tend to converge to a small set of canonical strategies,
whereas incorrect solutions diverge across many distinct error modes. Whether
the optimization direction induced by the vanilla contrastive signal aligns
with the verifier therefore depends on whether the target rollout $y$ and the
sampled negative hint $y_w^*$ share a comparable error type.
Table~\ref{tab:naive-failure-cases} enumerates the three resulting cases:
\begin{itemize}
    \item \textbf{Correct $y$.} The teacher endorses each token in $y$ under $y_c^*$ (driving $e_{c,t}$ up) and conflicts under $y_w^*$ (driving $e_{w,t}$ down), giving a strongly positive $e_{\text{ctr},t}$ that matches the verifier.
    \item \textbf{Incorrect $y$ with similar-error $y_w^*$.} Symmetrically, $e_{c,t}$ drops and $e_{w,t}$ rises, giving a strongly negative $e_{\text{ctr},t}$ that again matches the verifier.
    \item \textbf{Incorrect $y$ with dissimilar-error $y_w^*$.} $e_{c,t}$ still drops, but because $y_w^*$ commits a different mistake from $y$, the teacher does not endorse $y$'s tokens under it either; $e_{w,t}$ loses its expected positive sign, and $e_{\text{ctr},t}$ can carry either sign.
\end{itemize}
The third case is the failure mode: a substantial fraction of incorrect
rollouts end up with their tokens overall encouraged rather than penalized,
injecting substantial noise into the optimization. Actively selecting an
error-matched negative hint via online LLM-as-judge scoring is prohibitively
expensive at RL-training scale, so we instead marginalize over $K$
independently sampled negative hints, taking their probability-mean as the
negative branch. This spreads the teacher's mass across the error modes of
$\mathcal{G}^-$ rather than betting on a single one.

\begin{table}[htbp]
\caption{Behavior of the naive contrastive signal $e_{\text{ctr},t}^{(\text{Naive})}$ across three sampling cases. Arrows indicate the sign of each quantity: $\uparrow$ positive, $\downarrow$ negative. The first two cases yield a $e_{\text{ctr},t}$ that aligns with the verifier (\textcolor{cleanGreen}{\checkmark}), but the third case is the failure mode (\textcolor{naivered}{\ding{55}}): when $y_w^*$ does not share $y$'s error type, the negative branch $e_{w,t}$ no longer reliably endorses $y$'s tokens, and $e_{\text{ctr},t}$ loses its directional anchor.}
\centering
\small
\begin{tabular}{l l c c c}
\toprule
$y$ & $y_w^*$ relative to $y$ & $e_{c,t}$ & $e_{w,t}$ & $e_{\text{ctr},t}$ \\
\midrule
correct   & ---               & $\uparrow$  & $\downarrow$ & $\uparrow$ \quad\textcolor{cleanGreen}{\checkmark} \\
\midrule
incorrect & similar error     & $\downarrow$ & $\uparrow$  & $\downarrow$ \quad\textcolor{cleanGreen}{\checkmark} \\
incorrect & dissimilar error  & $\downarrow$ & $?$ & $?$ \quad\textcolor{naivered}{\ding{55}} \\
\bottomrule
\end{tabular}
\label{tab:naive-failure-cases}
\end{table}

\begin{figure}[t]
    \centering
    \includegraphics[width=1.0\linewidth]{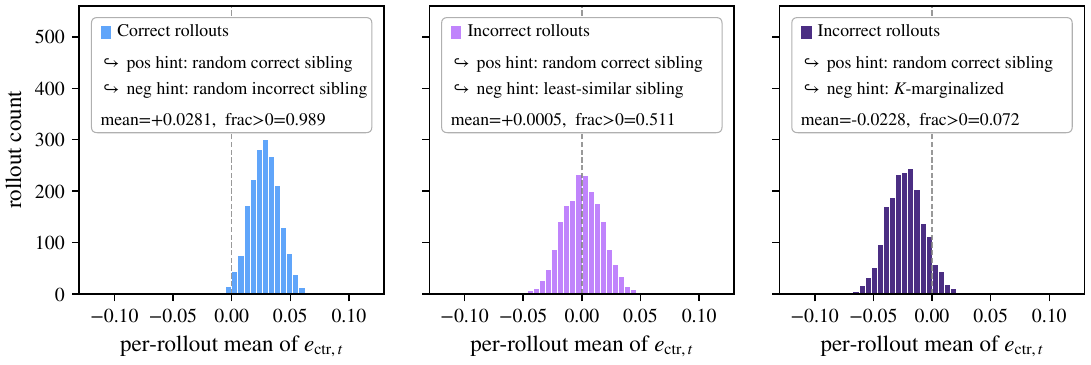}
    \caption{Rollout-level mean of $e_{\text{ctr},t}$, computed with one positive hint and one or more negative hints. The positive hint is a randomly sampled correct sibling in all three panels; the panels differ in the rollout type and the negative-hint strategy. \textbf{Left:} correct rollouts with a single randomly sampled negative hint, illustrating the expected positive signal of Case~1. \textbf{Middle:} incorrect rollouts with a single LLM-judged least-similar negative hint (the worst case, Case~3). \textbf{Right:} incorrect rollouts with $K$-marginalized negative hints.}
\label{fig:naive-failure-1}
\end{figure}

\paragraph{Empirical validation.}
Figure~\ref{fig:naive-failure-1} confirms the analysis through rollout-level
means of $e_{\text{ctr},t}$. Correct rollouts produce a cleanly positive signal
(left, frac~$>0$~=~0.989), validating Case~1. To isolate Case~3, the middle
panel uses the least-similar negative hint per incorrect $y$ (selected offline
by an LLM judge): the signal is centered at zero, with roughly half the
rollouts bearing the wrong sign. $K$-marginalization (right) restores a cleanly
negative-skewed signal, reducing the wrong-sign rate from $51\%$ to under $8\%$.

\subsection{Self-Referential Collapse and Excluding the Target Rollout}
\label{app:self-hint}

\paragraph{Self-conditioning induces over-confidence.}
Whenever a privileged hint is sampled from student rollouts, the target rollout $y$ itself may be selected as its own hint. Conditioning the teacher on the exact response being evaluated induces extreme over-confidence, making the resulting signal overly sharp and obscuring the token-level distinctions that $e_{\mathrm{ctr},t}$ is intended to capture.

\begin{figure}[t]
    \centering
    \includegraphics[width=1.0\linewidth]{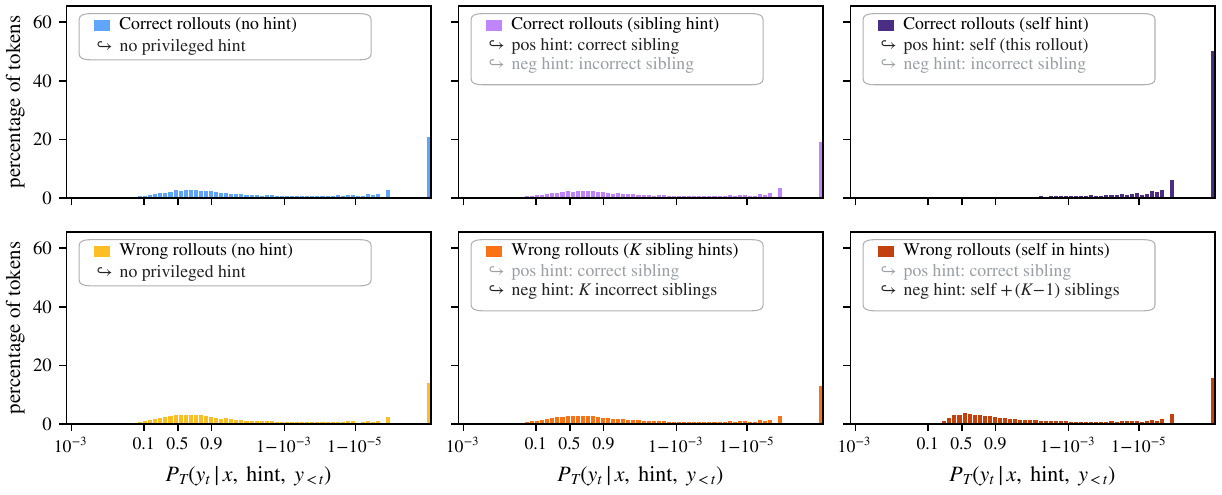}
    \caption{Distribution of teacher token probabilities $P_T(y_t \mid x, \text{hint}, y_{<t})$ under different hint sources, for correct (top) and incorrect (bottom) rollouts. Using a sibling hint leaves the distribution close to the no-hint case, whereas including the rollout itself as the hint shifts mass toward over-confident, high-probability tokens, motivating its exclusion from the hint pool.}
\label{fig:naive-failure-2}
\end{figure}

To characterize this effect, we compare conditioning on the target rollout itself with conditioning on another rollout for the same query, considering both correct and incorrect targets. As shown in Figure~\ref{fig:naive-failure-2}, for correct rollouts, a sibling hint leaves the overall distribution largely unchanged, whereas using the target itself as the hint raises the fraction of high-probability tokens ($P_T > 1-10^{-5}$) by roughly $30\%$. For incorrect rollouts, $K$-marginalization mitigates this effect, but including the target among the $K$ candidates still increases the high-probability fraction by about $10\%$.

We therefore exclude the target rollout $y$ from all rollout-derived hint pools. In our setting, this means drawing negative hints from $\mathcal{G}^-\setminus\{y\}$. More generally, if correct student rollouts are used as positive hints, the same exclusion should apply: positive hints should be drawn from $\mathcal{G}^+\setminus\{y\}$.

\section{Implementation Details}
\label{app:implement-detail}

This section presents the full implementation details that were not fully described in Section~\ref{sec:exp-setup} of the main text.

\subsection{Shared training setup}
\label{app:shared-hparams}

  All methods are implemented on top of the veRL~\citep{sheng2024hybridflow} PPO
  trainer with FSDP2 actor sharding and vLLM rollouts.
  All runs use full-parameter training (no LoRA) on $8\times$H20 GPUs.
  The settings in Table~\ref{tab:shared-hparams} are identical across
  GRPO, OPSD, SDPO, SRPO, RLSD, and RLCSD within a given (task, backbone)
  configuration. Only the per-step loss differs across methods.

  \begin{table}[h]
  \centering
  \small
  \caption{Hyperparameters shared by all six methods. }
  \label{tab:shared-hparams}
  \begin{tabular}{lll}
  \toprule
  Group & Hyperparameter & Value \\
  \midrule
  \multirow{5}{*}{Optimization}
    & Optimizer            & AdamW \\
    & Weight decay         & $0.01$ \\
    & LR warmup (linear)   & $50$ steps \\
    & LR   & $1 \times 10^{-6}$ \\
    & GPUs                 & $8\times$H20 \\
  \midrule
  \multirow{5}{*}{Batching}
    & Per-device batch     & $8$ \\
    & Group size $G$       & $8$ \\
    & PPO mini-batch       & $16$ \\
    & Rollout IS mode      & token-level \\
    & Rollout IS threshold & $2.0$ \\
  \midrule
  \multirow{4}{*}{Rollout sampling}
    & Temperature          & $1.0$ \\
    & Top-$p$ / Top-$k$    & $0.95\ /\ 20$ \\
    & Max prompt length    & $2{,}048$ \\
    & Max completion length & $16{,}384$ \\
  \midrule
  \multirow{5}{*}{Evaluation sampling}
    & Temperature                  & $0.6$ \\
    & Top-$p$ / Top-$k$            & $0.95\ /\ 20$ \\
    & Max completion length        & $38{,}912$ \\
    & Eval batch size              & $16$ \\
    & Samples per query            & $12$ (math, mean@12) / $1$ (K\&K, pass@1) \\
  \bottomrule
  \end{tabular}
  \end{table}

  \subsection{Method-specific hyperparameters}
  \label{app:method-hparams}

  Table~\ref{tab:method-hparams} lists the hyperparameters that vary
  across methods. We follow the \emph{original papers} wherever values are specified there.

\begin{table}[t]
  \centering
  \small
  \caption{Method-specific hyperparameters.}
  \label{tab:method-hparams}
  \resizebox{\textwidth}{!}{%
  \begin{tabular}{llll}
  \toprule
  Method & Hyperparameter & Value & Description \\
  \midrule
  \multirow{2}{*}{GRPO}
    & KL-to-ref coef.        & $10^{-3}$ & coef. on $\mathrm{KL}(\pi_\theta\|\pi_{\text{ref}})$. \\
    & PPO clip $\epsilon$    & $0.2$     & PPO-clip range. \\
  \midrule
  \multirow{7}{*}{OPSD}
    & KL-to-ref coef.        & $0$       & disabled. \\
    & JSD mixing $\beta$     & $0.0$     & $0$ = forward KL, $1$ = reverse KL. \\
    & Vocab clip             & $0.05$    & per-(token, vocab) divergence cap. \\
    & Full-logit distill     & yes       & dense top-$k$ + tail-bucket target. \\
    & Top-$k$                & $100$     & vocab indices per token. \\
    & Tail bucket            & yes       & residual mass outside top-$k$. \\
    & Teacher mode           & fixed     & frozen pretrained checkpoint. \\
  \midrule
  \multirow{7}{*}{SDPO\,/\,SRPO}
    & KL-to-ref coef.        & $0$       & disabled. \\
    & Mixing coef.\ $\alpha$ & $0.5$     & student/teacher mix in JSD target. \\
    & Full-logit distill     & yes       & dense top-$k$ + tail-bucket. \\
    & Top-$k$                & $100$     & vocab indices per token. \\
    & PPO clip ratio         & $2.0$     & IS-ratio clip on distill term. \\
    & EMA decay              & $0.95$    & EMA self-teacher decay. \\
    & Teacher mode           & EMA       & EMA of policy weights. \\
  \midrule
  \multirow{6}{*}{RLSD}
    & KL-to-ref coef.        & $0$       & disabled. \\
    & PPO clip $\epsilon$    & $0.2$     & outer PPO-clip on $\pi_\theta/\pi_{\text{old}}$. \\
    & Evidence clamp $\epsilon_w$ & $0.2$ & clamp on $w_t=\exp(\mathrm{sign}(A)\,(\log\pi_T\!-\!\log\pi_S))$. \\
    & Modulator $\lambda$    & $0.5$     & $\lambda{=}0$: GRPO; $\lambda{=}1$: full modulation. \\
    & $\lambda$ schedule     & $0.5\!\to\!0$ in $50$ steps & linear decay. \\
    & Teacher mode           & snapshot ($\tau_{\text{sync}}{=}10$) & hard copy every $10$ steps. \\
  \midrule
  \multirow{8}{*}{RLCSD (ours)}
    & KL-to-ref coef.        & $0$      & disabled. \\
    & PPO clip $\epsilon$    & $0.2$    & both paths of Eq.~\ref{eq:rlcsd-loss}. \\
    & Slope $\tau$           & $1.3$   & $\tanh$ temperature, Eq.~\ref{eq:rt}. \\
    & Scale $\lambda$        & $0.5$    & overall scale of $r_t$, Eq.~\ref{eq:rt}. \\
    & Mask threshold $\delta$ & $0.02$  & $m_t{=}\mathbb{1}[|r_t|{>}\delta]$, Eq.~\ref{eq:mt}. \\
    & Two-path weight $\eta$  & $0.5$   & modulated-path weight, Eq.~\ref{eq:rlcsd-loss}. \\
    & Neg.\ hints $K$         & $4$     & negative siblings averaged, Eq.~\ref{eq:final-ectr}. \\
    & Teacher mode            & fixed \\
  \bottomrule
  \end{tabular}}
  \end{table}

\section{Vocabulary Partitioning for Task / Style Statistics}
  \label{app:vocab-split}

  This section specifies the token partition of math reasoning tasks used
  in Figures~\ref{fig:ec-task-vs-style} and~\ref{fig:ctr-clean}.

  For each decoded token, we first strip the leading-space marker used by
  the Qwen3 tokenizer and lowercase the result. The token is then sent
  through the following rules in order, returning at the first match:

  \begin{enumerate}
  \itemsep0em
  \item[(1)] empty or whitespace-only $\rightarrow$ \textbf{style};
  \item[(2)] matches any of the math regexes
             (a digit \texttt{\textbackslash d};
              an arithmetic operator in
              \texttt{+\,-\,=\,*\,/\,<\,>\,×\,÷\,$\leq$\,$\geq$\,$\neq$};
              a LaTeX command \texttt{\textbackslash[A-Za-z]+};
              a double backslash;
              or one of \texttt{\$\,\textasciicircum\,\_})
             $\rightarrow$ \textbf{task};
  \item[(3)] the normalized form is in the math wordlist
             \texttt{\{mod, prime, factor, gcd, lcm, log, ln, sin, cos,
             tan, exp, integral, sqrt, boxed, frac, sum, prod, pi, alpha,
             beta, gamma, theta, delta, lambda, mu, sigma, infty, leq,
             geq, neq, cdot, times, div\}}
             $\rightarrow$ \textbf{task};
  \item[(4)] pure punctuation or a literal newline token
             ({\tt \textbackslash n}, {\tt \textbackslash\textbackslash n})
             $\rightarrow$ \textbf{style};
  \item[(5)] the normalized form is in the discourse wordlist
             (connectives \texttt{therefore, so, thus, hence, then, because,
             since}; hedges \texttt{wait, maybe, perhaps, seems, okay, ok,
             well, now, first, next, finally, actually, alternatively,
             however}; scaffolding \texttt{step, answer, let, lets};
             closed-class function words \texttt{is, are, us, we, the, a,
             an, of, to, for, in, on, by, at, as, and, or, but, if, yes,
             no, this, that, these, those, it, its, be, been, being, have,
             has, had, do, does, did, will, would, should,
             could, can, may})
             $\rightarrow$ \textbf{style};
  \item[(6)] otherwise $\rightarrow$ \textbf{neutral}.
  \end{enumerate}

\section{Computational Cost Analysis}

RLCSD introduces additional teacher-side computation compared with one-sided self-distillation methods, since it evaluates the teacher on both a correct hint and multiple wrong hints to form the contrastive signal. This extra cost is reflected in the \texttt{teacher log prob} time in Table~\ref{tab:wall_clock}: RLCSD spends \(14.210\)s per step on teacher scoring, compared with about \(9\)s for one-sided methods such as RLSD and OPSD.

However, this overhead is small compared with the dominant cost of autoregressive rollout generation. RLCSD spends \(473.89\)s per step on \texttt{gen}, so the extra teacher scoring is only a minor fraction of the generation cost. In other words, although RLCSD requires more teacher forwards, these forward-only evaluations add relatively little wall-clock overhead compared with sampling long reasoning trajectories.

More importantly, the end-to-end step time remains favorable. As shown in Table~\ref{tab:wall_clock}, RLCSD is the \emph{third fastest} method overall, behind only GRPO and RLSD. In particular, it is faster than dense-distillation baselines such as OPSD, SDPO, and SRPO. A likely reason is that these methods perform dense token-level distillation over the vocabulary during training, which makes the optimization stage itself much more expensive than the sampled-token modulation used in RLCSD and RLSD. 

\begin{table}[h]
\centering
\small
\caption{Average wall-clock time per training step (seconds) on math reasoning task, using Qwen3-4B as the base model. RLCSD spends more time on \texttt{teacher\_log\_prob} because it uses additional wrong-hint teacher forwards, but this overhead is small relative to rollout generation (\texttt{gen}). In total step time, RLCSD remains the third fastest method overall.}
\label{tab:wall_clock}
\begin{tabular}{lccc}
\toprule
Method & Gen (s/step) & Teacher log prob (s/step) & Total step (s/step) \\
\midrule
GRPO  & 503.78 & --     & 759.01  \\
OPSD  & 413.92 & 9.552  & 1035.54 \\
SDPO  & 495.64 & 9.701  & 961.14  \\
SRPO  & 411.01 & 9.566  & 1011.51 \\
RLSD  & 412.27 & 9.485  & 812.84  \\
RLCSD & 473.89 & 14.210 & 891.55  \\
\bottomrule
\end{tabular}
\end{table}

\section{Additional Results on Agentic Tasks}
\label{app:agentic}

\paragraph{Settings.} We evaluate Qwen3-4B on ALFWorld~\citep{shridhar2020alfworld} and WebShop~\citep{yao2022webshop}. We allow at most 30 interaction steps per episode on ALFWorld and 20 on WebShop, where each step consists of an agent action and the corresponding environment response. For both benchmarks, we enable thinking mode and impose a trajectory budget of 16,384 tokens, counting both model outputs and subsequent environment observations. At evaluation time, we sample one trajectory per task with a temperature of 0.6; we use top-$p$ of 1.0 without top-$k$ filtering for ALFWorld, and top-$p$ of 0.95 with top-$k$ of 20 for WebShop. For ALFWorld, we report overall success rate and success rates on the seen and unseen splits. For WebShop, we report the average reward, which includes partial credit and ranges from 0 to 1, and the success rate, defined as the fraction of episodes receiving a reward of 1. All success rates are reported as percentages.

\begin{table}[t]
\centering
\caption{Results on agentic tasks using Qwen3-4B. ALFWorld metrics are success rates (\%). WebShop Score is the average reward on a scale of $[0,1]$, and Success is the percentage of episodes receiving a reward of 1. Higher is better for all metrics. \textbf{Bold} marks the best result in each column, and \underline{underline} marks the second-best result.}
\label{tab:agentic}
\small
\setlength{\tabcolsep}{8pt}
\begin{tabular}{lccccc}
\toprule
& \multicolumn{3}{c}{\textbf{ALFWorld}}
& \multicolumn{2}{c}{\textbf{WebShop}} \\
\cmidrule(lr){2-4}\cmidrule(lr){5-6}
\textbf{Method}
& Overall (\%) & Seen (\%) & Unseen (\%)
& Score & Success (\%) \\
\midrule
Base  & 43.06 & 45.00 & 41.04 & 0.5081 & 18.4 \\
GRPO  & \underline{47.44} & 48.57 & 46.26 & 0.5170 & 18.8 \\
OPSD  & 46.35 & 42.85 & \textbf{50.00} & 0.5164 & 18.5 \\
SDPO  & 46.35 & \textbf{50.71} & 41.79 & \underline{0.5225} & 18.6 \\
SRPO  & 45.98 & 46.42 & 45.52 & 0.5194 & 18.7 \\
RLSD  & 45.62 & 43.57 & \underline{47.76} & 0.5218 & \underline{19.1} \\
\rowcolor{gray!30}
\textbf{RLCSD} & \textbf{48.17} & \underline{49.28} & 47.01 & \textbf{0.5300} & \textbf{19.4} \\
\bottomrule
\end{tabular}
\end{table}

\paragraph{Results.} As shown in Table~\ref{tab:agentic}, RLCSD achieves the highest overall success rate on ALFWorld and the best average reward and success rate on WebShop. On ALFWorld, RLCSD reaches 48.17\% overall success, outperforming the base model by 5.11 percentage points and the strongest baseline, GRPO, by 0.73 percentage points. Although SDPO and OPSD achieve the highest success rates on the seen and unseen splits, respectively, RLCSD performs consistently across both splits, yielding the best overall result. On WebShop, RLCSD achieves an average reward of 0.5300 and a success rate of 19.4\%, exceeding the strongest baselines for these metrics by 0.0075 and 0.3 percentage points, respectively. These results show consistent, albeit modest, improvements over the competing methods across both environments.

\section{Sensitivity Analysis of Hyperparameters in RLCSD}
\label{app:hyperparameter}

We investigate how RLCSD's hyperparameters affect token selection and training behavior using Qwen3-8B on DeepMath. The five hyperparameters are the number of negative hints $K$, the soft-normalization temperature $\tau$, the modulation scale $\lambda$, the selection threshold $\delta$, and the modulated-path weight $\eta$. Our default configuration is
\begin{equation}
(K,\tau,\lambda,\delta,\eta)=(4,1.3,0.5,0.02,0.5).
\end{equation}
We first analyze their direct effects on fixed contrastive signals, then examine the default training dynamics and controlled comparisons in which one hyperparameter is varied at a time.

\subsection{How the Hyperparameters Affect the Learning Signal}

\paragraph{Token selection depends on a joint effective threshold.}
RLCSD transforms the raw contrastive signal into a bounded modulation and selects tokens according to
\begin{equation}
r_t=\lambda\tanh\left(\frac{e_{\mathrm{ctr},t}}{\tau}\right),
\qquad
m_t=\mathbf{1}(|r_t|>\delta).
\end{equation}
For $\tau>0$ and $0<\delta<\lambda$, the selection criterion is equivalent to
\begin{equation}
m_t=\mathbf{1}\left(|e_{\mathrm{ctr},t}|>h_{\mathrm{eff}}\right),
\qquad
h_{\mathrm{eff}}
=\tau\operatorname{arctanh}\left(\frac{\delta}{\lambda}\right).
\end{equation}
Thus, $\tau$, $\lambda$, and $\delta$ jointly determine selection coverage. On fixed raw signals, increasing $\tau$ or $\delta$ raises the effective threshold and selects fewer tokens, whereas increasing $\lambda$ lowers it and selects more tokens. The default configuration corresponds to $h_{\mathrm{eff}}\approx0.0520$.

Table~\ref{tab:hyperparameter-mechanism} summarizes these analytical relationships. Different configurations can produce identical selection masks while assigning different modulation magnitudes. For example, doubling $\lambda$ and halving $\delta$, considered separately from the same default configuration, produce the same effective threshold at fixed $\tau$. However, only the former scales the modulation values.

\begin{table}[t]
\centering
\caption{Analytical effects of changing one hyperparameter from the default configuration. The effective threshold $h_{\mathrm{eff}}$ and local gain $\lambda/\tau$ are calculated from the modulation formula. Selection trends assume fixed raw contrastive signals and describe non-strict changes relative to the default configuration.}
\label{tab:hyperparameter-mechanism}
\small
\setlength{\tabcolsep}{7pt}
\begin{tabular}{lcccc}
\toprule
Configuration
& $h_{\mathrm{eff}}$
& Local gain $\lambda/\tau$
& Bound $\lambda$
& Selection coverage \\
\midrule
Default          & 0.0520 & 0.3846 & 0.50 & --- \\
\midrule
$\tau=0.65$      & 0.0260 & 0.7692 & 0.50 & Increases \\
$\tau=2.6$       & 0.1041 & 0.1923 & 0.50 & Decreases \\
\midrule
$\lambda=0.25$   & 0.1042 & 0.1923 & 0.25 & Decreases \\
$\lambda=1.0$    & 0.0260 & 0.7692 & 1.00 & Increases \\
\midrule
$\delta=0.01$    & 0.0260 & 0.3846 & 0.50 & Increases \\
$\delta=0.04$    & 0.1042 & 0.3846 & 0.50 & Decreases \\
\midrule
$\eta=0.25$      & 0.0520 & 0.3846 & 0.50 & Unchanged \\
$\eta=1.0$       & 0.0520 & 0.3846 & 0.50 & Unchanged \\
\bottomrule
\end{tabular}
\end{table}

\paragraph{Modulation strength differs from selection coverage.}
For small $|e_{\mathrm{ctr},t}|/\tau$, the modulation is approximately linear:
\begin{equation}
r_t\approx\frac{\lambda}{\tau}e_{\mathrm{ctr},t}.
\end{equation}
The ratio $\lambda/\tau$ controls the local gain, while $\lambda$ bounds the modulation magnitude. Increasing $\lambda$ scales every modulation value proportionally. Decreasing $\tau$ also strengthens the modulation, but pushes more tokens toward saturation and compresses differences among large raw signals.

In contrast, changing $\delta$ leaves every $r_t$ unchanged and affects only selection. A larger $\delta$ therefore retains fewer, stronger signals without changing the average modulation magnitude over all tokens. For $\lambda$ and $\tau$, changes in modulation magnitude and selected-set membership occur together, so the average magnitude among selected tokens need not follow the same trend as the all-token average. The sign-preserving clamp further limits the effective advantage change when a modulation would otherwise reverse the outcome-based update direction.

\paragraph{Negative-hint marginalization controls estimation variability.}
The parameter $K$ acts upstream by changing the negative-branch estimate:
\begin{equation}
e_{\mathrm{ctr},t}
=
\log p_{c,t}
-
\log\left(\frac{1}{K}\sum_{k=1}^{K}p_{w,k,t}\right),
\end{equation}
where $p_{c,t}$ and $p_{w,k,t}$ denote teacher probabilities under the correct and incorrect references, respectively. Under independent sampling, the variance of the negative-branch probability mean decreases as $1/K$. Increasing $K$ can therefore reduce sensitivity to individual negative hints and improve coverage of the error modes represented in the hint pool. However, it does not imply a monotonic change in the absolute contrastive signal or the selected-token ratio, because both depend on the sampled probabilities and the subsequent logarithm.

\paragraph{Path weighting interacts with selection sparsity.}
For a rollout with nonempty unmodulated and modulated token sets, the objective is
\begin{equation}
J
=
\frac{1}{|\mathcal U|}
\sum_{t\in\mathcal U}\psi_t(A_{\mathrm{ORM}})
+
\frac{\eta}{|\mathcal M|}
\sum_{t\in\mathcal M}\psi_t(\widetilde A_t).
\end{equation}
Writing $q=|\mathcal M|/(|\mathcal U|+|\mathcal M|)$, the ratio of per-token normalization weights is
\begin{equation}
\frac{\eta/|\mathcal M|}{1/|\mathcal U|}
=
\eta\frac{1-q}{q}.
\end{equation}
Selecting fewer tokens therefore concentrates the modulated-path weight on a smaller subset rather than proportionally reducing its nominal weight. At $\eta=0.5$, this ratio is $2.0$ for $q=0.2$ and approximately $1.17$ for $q=0.3$, before accounting for advantage magnitudes and PPO clipping.

Unlike $\lambda$, $\eta$ does not directly change the signal or the selection mask. It scales the entire modulated path, including the outcome-level anchor within $\widetilde A_t$. During training, however, changing $\eta$ can indirectly alter subsequent selection statistics by changing the policy and its rollout distribution.

\subsection{Training Dynamics under the Default Configuration}

Table~\ref{tab:default-dynamics} reports the default configuration's training statistics across three windows. The selected-token ratio remains close to $28\%$--$30\%$, decreasing from $30.01\%$ over steps 1--50 to $28.24\%$ over steps 351--400. Response length is maintained, changing from approximately $9{,}423$ to $9{,}617$ tokens, while training reward remains comparable across the middle and late stages. Actor entropy decreases from $0.321$ to $0.183$ without an accompanying collapse in response length. These observations characterize the default configuration as maintaining selective modulation throughout training while preserving extended responses.

\begin{table}[t]
\centering
\caption{Training dynamics of Qwen3-8B on DeepMath under the default RLCSD configuration. Each entry is the mean of the logged per-step metric over the indicated training window.}
\label{tab:default-dynamics}
\small
\setlength{\tabcolsep}{10pt}
\begin{tabular}{lccc}
\toprule
Metric & Steps 1--50 & Steps 201--250 & Steps 351--400 \\
\midrule
Selected tokens (\%) & 30.01 & 28.62 & 28.24 \\
Response length     & 9423  & 9346  & 9617  \\
Training reward     & 0.736 & 0.753 & 0.749 \\
Actor entropy       & 0.321 & 0.201 & 0.183 \\
\bottomrule
\end{tabular}
\end{table}

\subsection{Controlled Hyperparameter Comparisons}

\paragraph{Evaluation protocol.}
We vary one hyperparameter at a time while keeping the remaining settings and training budget fixed. We report the selected-token ratio, response length, training reward, and actor entropy averaged over steps 351--400, together with AIME25 mean@12 at step 400. These training-level measurements capture both the direct effects described above and the indirect effects of changes in the learned policy.

\begin{table}[t]
\centering
\caption{Hyperparameter sensitivity of RLCSD on Qwen3-8B and DeepMath. Each configuration changes one parameter from the default setting. Training statistics are averaged over steps 351--400; AIME25 mean@12 is evaluated at step 400.}
\label{tab:hyperparameter-training}
\small
\resizebox{0.8\textwidth}{!}{%
\begin{tabular}{lccccc}
\toprule
Configuration
& Selected (\%)
& Length (tokens)
& Reward
& Entropy
& AIME25 (\%) \\
\midrule
Default             & 28.24 & 9617  & 0.749 & 0.183 & 69.72 \\
\midrule
$K=1$               & 32.67 & 9148  & 0.731 & 0.207 & 66.94 \\
$K=2$               & 29.83 & 9486  & 0.743 & 0.192 & 68.89 \\
\midrule
$\tau=0.65$         & 39.56 & 10083 & 0.744 & 0.216 & 68.61 \\
$\tau=2.6$          & 18.72 & 9324  & 0.738 & 0.174 & 68.06 \\
\midrule
$\lambda=0.25$      & 19.15 & 9408  & 0.741 & 0.178 & 68.33 \\
$\lambda=1.0$       & 40.38 & 10426 & 0.736 & 0.231 & 67.78 \\
\midrule
$\delta=0.01$       & 38.91 & 9702  & 0.747 & 0.189 & 69.17 \\
$\delta=0.04$       & 18.46 & 9538  & 0.750 & 0.181 & 70.00 \\
\midrule
$\eta=0.25$         & 27.63 & 9361  & 0.740 & 0.176 & 68.61 \\
$\eta=1.0$          & 29.47 & 10054 & 0.745 & 0.204 & 69.44 \\
\bottomrule
\end{tabular}%
}
\end{table}

\paragraph{Selection coverage follows the effective-threshold analysis.}
The changes in selected-token ratio agree with the predicted directions. Decreasing $\tau$ to $0.65$, increasing $\lambda$ to $1.0$, or decreasing $\delta$ to $0.01$ expands coverage to approximately $39\%$--$40\%$. The opposite changes reduce coverage to approximately $18\%$--$19\%$. Although these relationships are derived for fixed raw signals, their directions remain evident after training. Across these configurations, AIME25 ranges from $67.78\%$ to $70.00\%$, indicating that substantial changes in selection coverage do not translate into proportional changes in validation performance.

\paragraph{Multiple negative hints improve performance.}
Increasing $K$ from 1 to 2 and 4 improves AIME25 from $66.94\%$ to $68.89\%$ and $69.72\%$, respectively. Training reward also increases from $0.731$ to $0.743$ and $0.749$. These results are consistent with the benefit of a more reliable negative-branch estimate. The incremental accuracy gain decreases from $1.95$ points for $K=1$ to $K=2$ to $0.83$ points for $K=2$ to $K=4$, supporting the use of multiple hints while suggesting diminishing gains within the tested range.

\paragraph{Stronger modulation does not necessarily improve accuracy.}
Increasing $\lambda$ to $1.0$ or decreasing $\tau$ to $0.65$ produces longer responses than the default configuration, reaching $10{,}426$ and $10{,}083$ tokens on average, respectively. However, their AIME25 scores decrease to $67.78\%$ and $68.61\%$. Weaker modulation also yields lower scores: $\lambda=0.25$ and $\tau=2.6$ achieve $68.33\%$ and $68.06\%$. These comparisons support balancing modulation strength rather than maximizing it, and show that longer responses alone do not guarantee better reasoning performance.

\paragraph{Sparser selection can preserve performance.}
Increasing $\delta$ to $0.04$ reduces the selected-token ratio from $28.24\%$ to $18.46\%$ while maintaining comparable AIME25 performance ($70.00\%$ versus $69.72\%$). Decreasing $\delta$ to $0.01$ expands coverage to $38.91\%$ but yields $69.17\%$. These results suggest that broader coverage is not always necessary: independent normalization allows a smaller selected set to retain substantial influence on optimization. The small accuracy difference between $\delta=0.04$ and the default does not establish a clear preference between them.

\paragraph{Modulated-path weighting has a moderate effect in the tested range.}
Changing $\eta$ to $0.25$ or $1.0$ yields AIME25 scores of $68.61\%$ and $69.44\%$, respectively, compared with $69.72\%$ for the default. The selected-token ratio remains relatively close to the default, ranging from $27.63\%$ to $29.47\%$, consistent with $\eta$ affecting selection only indirectly through training. These comparisons support $\eta=0.5$ as a reasonable operating point, while the performance differences do not indicate that it is uniquely optimal.